\documentclass[11pt]{article}
\usepackage{float}

\usepackage[preprint]{acl}

\usepackage{times}
\usepackage{latexsym}
\usepackage{amsmath}

\usepackage[T1]{fontenc}

\usepackage[utf8]{inputenc}

\usepackage{microtype}

\usepackage{inconsolata}

\usepackage{graphicx}

\usepackage{xcolor}

\usepackage{multirow}
\usepackage{booktabs}
\usepackage{makecell}
\usepackage{pdflscape}

\usepackage{CJKutf8}

\usepackage{subcaption}
\usepackage[plain]{fancyref}

\title{Language Bias under Conflicting Information in Multilingual LLMs}

\author{
         Robert \"{O}stling \\ Stockholm University \\  Sweden \\ \texttt{robert@ling.su.se}
 \And
         Murathan Kurfal{\i} \\ RISE Research Institutes of Sweden \\  Sweden \\ \texttt{murathan.kurfali@ri.se}
         }

\begin{document}
\maketitle
\begin{abstract}
Large Language Models (LLMs) have been shown to contain biases in the process of integrating conflicting information when answering questions. Here we ask whether such biases also exist with respect to which language is used for each conflicting piece of information. To answer this question, we extend the conflicting needles in a haystack paradigm to a multilingual setting and perform a comprehensive set of evaluations with naturalistic news domain data in five different languages, for a range of multilingual LLMs of different sizes. We find that all LLMs tested, including GPT-5.2, ignore the conflict and confidently assert only one of the possible answers in the large majority of cases. Furthermore, there is a consistent bias across models and prompting languages in which languages are preferred, with a general bias against Russian and, for the longest context lengths, in favor of Chinese. The language preferences are consistent between models trained inside and outside of mainland China, though somewhat stronger in the former category. There is also a general tendency among models to prioritize information that matches the language used for prompting. We hope to make users and developers of multilingual LLMs aware of this category of biases, to spur further research on their causes and possible mitigation.
\end{abstract}

\section{Introduction}

Given the enormous impact and real-world deployment of large language models (LLMs) in recent years, it is essential to evaluate how these models deal with the complexity of information that they are used to process. One particular such complexity is contradictory information, which is abundant in today's information landscape. A very common application for LLMs is to summarize information from multiple sources, and if sources disagree on some points, it is of critical importance to know how these disagreements are reflected in the output. Previous work \citep{kurfali-2025-conflicting} has investigated this question with respect to factors such as repetition and positioning of information in the context window given to models, by using a \emph{conflicting needles in a haystack} paradigm where conflicting pieces of information (\emph{needles}) are inserted into large amounts of unrelated textual information (\emph{haystacks}). We expand on this line of research by systematically varying the \emph{languages} of the needles and haystacks in order to test whether the language information is presented in affects its chances of making it to the final output.

Specifically, we ask the following research questions:
\begin{itemize}
    \item RQ1: How do current multilingual LLMs behave when faced with conflicting information in a naturalistic information retrieval setting?
    \item RQ2: Are current multilingual LLMs biased with respect to which language they prioritize when extracting contradictory information?
    \item RQ3: Are any languages consistently favored or disfavored across LLMs?
    \item RQ4: Does the country of origin of the LLMs influence their handling of contradictory information?
\end{itemize}

We identify four major contributions of this paper, both methodological (1) and empirical on the (in)ability to deal with conflicting information (2) as well as a detailed look at how the biases made apparent by this failure are structured (3, 4).
\begin{enumerate}
    \item We introduce the \emph{multilingual conflicting needles in a haystack} paradigm for investigating language biases in multilingual LLMs, using manually constructed naturalistic templates that simulate a real-world information retrieval scenario.
    \item Contrary to our expectation, we find that even the state-of-the-art models, such as GPT-5.2, consistently fail to detect conflicting information, even under the easiest conditions in our experiments.
    \item We find that there is a clear language bias as to which of two contradictory pieces of information is chosen by a model, and that this bias is surprisingly consistent across models.
    \item We find that the patterns of bias are similar between models trained in the West (North America and Europe) as compared to the East (PRC), but that there seems to be a slight bias against English in PRC-trained models only.
\end{enumerate}

\section{Related Work}

The Needle In A Haystack (NIAH) paradigm, originally introduced by \citet{kamradt2023needle}, has been used in LLM research to test information retrieval under controlled circumstances. The key idea is to build a collection of text (``haystack'') into which one or more pieces of information (``needles'') are inserted. An LLM is then asked to retrieve the information, and the rate of success indicates how well the LLM under test is able to deal with this particular configuration of haystack and needles.
\citet{hsieh2024rulerwhatsrealcontext} compared retrieval performance with contexts of different lengths, with the goal of mapping how usable context size compares to the architectural context size limit for common LLMs. Given their finding that many models show a substantial drop in retrieval performance with longer contexts, we systematically vary context length in our experiments.
\citet{Wang2024multimodalhaystack} extended the NIAH paradigm to multimodal data, using combined textual (English) and visual haystacks, finding that the accuracy for retrieving, counting, and reasoning from visual information was considerably lower than from information in text.

\citet{kurfali-2025-conflicting} extended the NIAH paradigm to \emph{conflicting} needles, where contradictory information is inserted into a haystack of text. They found that LLMs in general do not identify the conflict, and that in choosing which needle to retrieve, there are both model-specific biases and biases that are consistent across models. In particular, models tend to rely more on repeated needles. \citet{schuster2026factswinllmsource} extended this research to study how LLMs take source credibility into account when faced with conflicting information. They find that repetition of information can even make LLMs prioritize information from unknown or unreliable sources over high-credibility sources. While they were unable to mitigate this repetition bias with prompting-based approaches alone, they propose a fine-tuning method which reduces repetition bias and increases reliance on source credibility.
Other mitigation approaches include training models to explicitly report sources for their claims \citep{shaier-etal-2024-adaptive}.

\citet{lovering2025findinginconsistenciesdocuments} also studied how LLMs deal with conflicting information, both in a conflicting NIAH scenario with inserted needles, and by identifying naturally occurring contradictions within documents. They found that top models can identify about 60\% of contradictory needles and achieve a precision of about 50\% when identifying naturally occurring contradictions. Similar results have been found in other studies focusing on direct detection of contradictions \citep{li-etal-2024-contradoc,tan2026improvedevidenceextractiondocument}. In general, this setting leads to the identification of a larger share of contradictions than the other studies referred to above \citep{kurfali-2025-conflicting,schuster2026factswinllmsource}, but their focus is on using state-of-the-art models with chain-of-thought (CoT), and prompting with the explicit goal of looking for contradictions. This contrasts with our focus, which is to test how contradictions are handled during ``mundane'' tasks like information retrieval.

Given the different political systems, level of information control, and general cultural differences between mainland China (PRC) and what could be loosely referred to as the ``west'' (represented in our work by the US, Canada, and EU), it is reasonable to ask to what extent LLMs from these two areas are different. Previous work has shown that, for instance, the dominating political ideology in a country has an effect on the sentiments expressed by LLMs about prominent people \citep{buyl2026large}. We are interested in whether there are differences with respect to how different languages are treated. \citet{wenyi2025chinesechineselanguagemodels} found that in spite of what one may expect given the political and cultural differences, performance of PRC-trained LLMs is mostly in line with that of ``western'' models on a broad sample of languages including minority languages of the PRC, Mandarin, and several major languages from Asia and Europe. We test (RQ4) whether this result also holds for the more subtle biases introduced by conflicting multilingual information.

\section{Method}

We follow \citet{kurfali-2025-conflicting} in inserting multiple conflicting \emph{needles}, artificially generated pseudo-facts, into \emph{haystacks} of authentic factual news text. 
Here we extend previous work by using \emph{multilingual} haystacks, which allows us to investigate how multilingual LLMs select information depending on which language it is presented in.

A \emph{haystack} in this work consists of up to $N = 34$ news articles of approximately 25~000 English words in total, originally in English [eng] but machine-translated\footnote{We initially used the translation model of \citet{cui-etal-2025-multilingual}, but found the translation quality was low for some of the languages, so used Google Translate for the results presented in this work.} into Chinese [cmn], German [deu], Russian [rus] and Turkish [tur]. Simplified characters are used for Chinese, Cyrillic letters for Russian, and Latin letters for the remaining languages. In order to simulate a realistic information retrieval scenario from a multilingual collection of news articles, we construct our haystacks using articles from established news outlets. To facilitate reproducibility, we only source articles from media outlets\footnote{We downloaded articles from Wikinews (CC-BY) and ProPublica (CC-NC-ND), as well as transcripts from Democracy Now! (CC-NC-ND)} that publish content under Creative Commons licenses.

In our work, a specific haystack contains two languages, $L_1$ and $L_2$ (typically different). The two articles into which the needles are inserted are tied to $L_1$ and $L_2$, in that order, but the languages for each of the remaining $N-2$ articles are chosen independently and pseudo-randomly from a uniform (50/50) distribution over $L_1$ and $L_2$. The permutation of the $N$ articles is also pseudo-randomly chosen. In both cases, the pseudo-random seed is based on the needle to be inserted, which ensures that language assignments and permutations are comparable across languages for a given needle. \Fref{tab:haystack-configuration} summarizes the variables used to determine a haystack.

We define four categories of needles. Each category has two selected articles, chosen manually to topically fit with the theme of the needle. For instance, the articles selected for category 1 needles are both about the entertainment industry. For this purpose, we only use articles with a permissive license (CC-BY) which permits rewriting the article. 
\Fref{fig:contrastive-example} gives an example in context, and \Fref{tab:needles} lists the English versions of the eight needles we use. Data including translations into all languages are available in the Supplementary Materials, except for CC-NC-ND licensed articles where translation scripts are provided as translations could be considered derivative work.

\begin{table}[tb]
    \centering
    \begin{tabular}{lp{0.7\linewidth}}
        Variables & Values \\
        \hline
        Category & 1, 2, 3, 4 \\
        $L_1$, $L_2$ & cmn, deu, eng, rus, tur \\
        $X_1$, $X_2$ & Delcroft, Quellman, Pikehart \\
        $Y$ & Cinderfax, Noiseweld, Motelvine, Brovencia, Clevantra, Teraluxis  \\
    \end{tabular}
    \caption{Configuration for a given haystack. The name of the entity $Y$ as well as the language assignment (of the non-needle articles) and order of articles is chosen deterministically in a pseudo-random manner, conditioned only on the question category and pair of surnames. This implies that each pair $L_1$, $L_2$ of languages has a matching haystack configuration in all the other language pairs. Names were generated by GPT-5 and manually verified by web searches to be very rare or unused.}
    \label{tab:haystack-configuration}
\end{table}

\begin{table*}[tb]
    \centering
    \begin{tabular}{llp{0.75\linewidth}}
         Cat. & Article & Template \\
         \hline
         1 & \texttt{wn250819}/3 & John SURNAME, the original lead vocalist of BANDNAME, praised Rondell's work on the album cover picture. \\
         1 & \texttt{wn250924}/4 & The original lead vocalist of BANDNAME, John SURNAME, called the early death of Brett James a tragic loss for country music in the United States. \\
         \hline
         2 & \texttt{wn250830}/1 & Paul SURNAME, the editor of PAPERNAME, called the events a tragic story. \\
         2 & \texttt{wn250806}/5 & The editor of PAPERNAME, Paul SURNAME, called the accident highly entertaining. \\
         \hline
         3 & \texttt{wn250918}/7 & The long-time CEO of COMPANYNAME, George SURNAME, disagreed with the court's decision. \\
         3 & \texttt{wn250910}/5 & United States businessman George SURNAME, CEO of COMPANYNAME, witnessed the incident and said he is shocked by it. \\
         \hline
         4 & \texttt{wn250724}/5 & Richard SURNAME, chairman of the ORGNAME organization, expects Mitchell to resign from office soon. \\
         4 & \texttt{wn250723}/5 & The chairman of the ORGNAME organization, Richard SURNAME, supported the government's plan. \\
    \end{tabular}
    \caption{English needle templates used in this work, along with their category and the article/paragraph number in which they are inserted. The same eight needles have also been translated into the other four languages.}
    \label{tab:needles}
\end{table*}

An LLM is given one haystack as context, and inside the haystack we insert two sentences at different locations with the same value of Y but (typically) distinct $X_1$ and $X_2$. By querying an LLM for the lead vocalist/editor/CEO/chairman of Y, we can see which of $X_1$ and/or $X_2$ is retrieved.
The main novelty in this work is our use of \emph{multilingual} haystacks, which allows us to investigate to what extent the language of the needle and the language of the prompt influence which $X$ is retrieved. In particular, if swapping the languages that $X_1$ and $X_2$ are presented in results in the opposite $X$ being retrieved, we interpret that as indicating a bias towards the language that the retrieved $X$ was presented in. \Fref{fig:contrastive-example} illustrates one pair of English/Chinese contrastive haystacks. If haystack (a) yields ``Ashwren'' while haystack (b) yields ``Delcroft'' when a given LLM is queried for the name of the original lead vocalist of Cinderfax, this indicates a possible bias towards Chinese in this model.

We automatically generate 900 haystacks for each model, by producing all combinations of pseudo-fact category (4), ordered pairs of surnames ($3^2 = 9$, of which 6 are conflicting), ordered pairs of languages ($6\cdot5/2 = 15$, of which 5 are monolingual and 10 are bilingual). Each haystack is queried using each of the five prompt languages, for a total of 4500 LLM queries per model. The 900 haystacks can be categorized as follows:
\begin{itemize}
    \itemsep0em 
    \item 480 bilingual conflicting haystacks (4 questions times 6 conflicting needles times 10 language pairs times two language pair directions) which form the basis of our analysis, after dividing them into 240 contrastive pairs.
    \item 120 monolingual conflicting haystacks ($4 \cdot 6 \cdot 5$) which we only use as part of the analysis of conflict detection. 
    \item 240 bilingual non-conflicting haystacks ($4 \cdot 6 \cdot 10 \cdot 2$) which we do not analyze further in this work.
    \item 60 monolingual non-conflicting haystacks ($4 \cdot 3 \cdot 5$) which we only use as a control to verify that models are able to perform the retrieval task under optimal conditions. 
\end{itemize}

To answer the question of whether a given language is preferred over another by a model, we treat the 240 contrastive pairs as independent experiments with four possible outcomes:
\begin{enumerate}
    \itemsep0em 
    \item The surname $X$ presented in language $L_1$ is retrieved in both haystacks.
    \item The surname $X$ presented in language $L_2$ is retrieved in both haystacks.
    \item The same surname $X$ is retrieved in both haystacks of the pair, that is, the choice was based on factors other than language.
    \item Retrieval failure: neither of $X_1$ or $X_2$ is retrieved in one or both of the haystacks.
\end{enumerate}
Discarding the last two cases, we count the number of instances of (1) and (2) and use a binomial null model with $p = 0.5$, to account for the possibility of switching languages sometimes introducing non-systematic changes of which language's needle being retrieved, as well as inherent non-deterministic behavior during model inference. If (1) or (2) occurs significantly more often than the null model predicts, we consider this evidence of $L_1$ being preferred over $L_2$ (1), or vice versa (2).

\begin{figure}[t]
    \begin{subfigure}{\linewidth}
    \begin{quote}
    \small
    [...]
    \textcolor{lightgray}{It took a long time to persuade Ronnie to stand exactly as I wanted but in the end he was very brave and it was a perfect composition."}
    John \textbf{Delcroft}, the original lead vocalist of \textbf{Cinderfax}, praised Rondell's work on the album cover picture.
    \textcolor{lightgray}{He sustained numerous injuries throughout his career, including but not limited to concussions and broken ribs.} \newline
    [...] \newline
    \begin{CJK*}{UTF8}{gbsn}
    \textcolor{lightgray}{Jesus Take the Wheel 荣获格莱美“最佳乡村歌曲”奖。它还被提名为“年度歌曲”。}
    \textbf{Cinderfax}的原主唱John \textbf{Ashwren}称Brett James的早逝是美国乡村音乐的悲剧性损失。
    \end{CJK*}
    [...]
    \end{quote}
    \caption{$L_1 = \textrm{English}$, $L_2 = \textrm{Chinese}$}
    \end{subfigure}

    \vspace{1em}

    \begin{subfigure}{\linewidth}
    \begin{quote}
    \small
    [...]
    \begin{CJK*}{UTF8}{gbsn}
    \textcolor{lightgray}{我花了很长时间才说服罗尼完全按照我想要的方式站着，但最终他非常勇敢，这是一张完美的构图。”}
    \textbf{Cinderfax}乐队原主唱John \textbf{Delcroft}对Rondell的专辑封面图片创作工作表示赞赏。
    \textcolor{lightgray}{他在整个职业生涯中遭受过多次伤害，包括但不限于脑震荡和肋骨骨折。}
    \end{CJK*}
    \newline
    [...] \newline
    \textcolor{lightgray}{Jesus Take the Wheel won a Grammy award for "Best Country Song." It was also nominated for "Song of the Year."}
    The original lead vocalist of \textbf{Cinderfax}, John \textbf{Ashwren}, called the early death of Brett James a tragic loss for country music in the United States.
    [...]
    \end{quote}
    \caption{$L_1 = \textrm{Chinese}$, $L_2 = \textrm{English}$}
    \end{subfigure}

    \caption{Excerpts from a pair of contrastive haystacks with $Y = \textrm{Cinderfax}$, $X_1 = \textrm{Delcroft}$, $X_2 = \textrm{Ashwren}$. Each haystack contains up to about 25~000 words. The contents of (a) and (b) are identical except that the languages are swapped for the two needle sentences and the article they are integrated to. In the excerpts, only parts of two articles are shown. The remaining articles are in the same order and language across the two haystacks. Light gray indicates authentic news text that is part of the context (most of which is left out for brevity), while the text in black are generated from our manually constructed needle templates described in \Fref{tab:needles}.}
    \label{fig:contrastive-example}
\end{figure}

\section{Experimental Setting}

We performed the experiments on a set of 12 LLMs (listed in \Fref{tab:outcomes} and described in more detail in Appendix~\ref{sec:models}).
We used greedy decoding or a sampling temperature of 0 when applicable, aiming at getting as close to a deterministic results as possible given existing infrastructure. For the OpenAI models, we used their API, and for all other models we ran the experiments on a single H100 GPU. In total, we used approximately 100 GPU hours and USD 331.53 of API credits. For chain-of-thought models, we used the lowest setting of ``thinking effort''. For chat models, we constructed a single long ``user'' message consisting of the haystack, a blank line, and the question. For non-chat models, we used the simple template ``Question: [question text]'', followed by a blank line and ``Answer:''.

Code and full results data files are provided as Supplementary Materials.

\section{Results and Discussion}
\label{sec:res}

\subsection{RQ1: handling of conflicting information}

Starting with our first research question, we begin by looking at how the LLMs we study are able to deal with conflicting information overall. \Fref{tab:outcomes} shows that all LLMs are unable to reliably identify conflicting information in the haystacks.\footnote{Due to the large number of queries, we classify the outputs automatically using the heuristic that if both valid answers are found in the output, this likely represents a refusal due to identifying the conflict. While in theory it is possible for a model to correctly mention the existence of a conflict without specifying the alternatives, during our manual inspection of LLM outputs we have so far not found any exceptions to this heuristic.} For small haystacks (approximately 1~000 words) almost all queries result in a single answer without mentioning the alternative. With large haystacks (approximately 25~000 words) the main qualitative change is that, with some models, a larger proportion of queries fail to give \emph{any} of the two correct answers. Results for monolingual haystacks are found in Appendix~\ref{sec:monolingual-haystacks} (\Fref{tab:outcomes-monolingual}), and are qualitatively very similar.

Our prompt specifies the expected output format by adding the sentence ``Answer with only the full name.'' This decision was made to simulate a typical prompting-based information retrieval scenario, where the user does not want any additional information besides the name. In order to test whether this sentence restricts the model too much, we also repeated our experiments for some settings with the sentence removed, leaving the task more open. Although the number of detected contradictions increases, as expected, we still observe the same qualitative pattern: a large majority of conflicting needles go undetected. A limited set of experiments with this reduced prompt are summarized in Appendix~\ref{app:outcomes-short-prompt}, but apart from this we refrain from further prompt exploration due to the associated computational cost.

\begin{table*}[t]
\centering
\begin{tabular}{l|rrr|rrr}
 & \multicolumn{3}{c}{1~000 words} & \multicolumn{3}{c}{25~000 words} \\
Model & Both & None & One & Both & None & One \\
\hline
\textsc{gemma-3-27b-it} & 14 & 0 & 2386 & 0 & 178 & 2222 \\
\textsc{gemma-3-4b-it} & 22 & 22 & 2356 & 0 & 676 & 1724 \\
\textsc{Llama-3.1-8B-Instruct} & 9 & 49 & 2342 & 0 & 41 & 2359 \\
\textsc{gpt-5.2-2025-12-11} & 0 & 0 & 2400 & 13 & 2 & 2385 \\
\textsc{gpt-5-mini-2025-08-07} & 23 & 0 & 2377 & 12 & 0 & 2388 \\
\textsc{gpt-5-nano-2025-08-07} & 1 & 2 & 2397 & 0 & 6 & 2394 \\
\textsc{c4ai-command-r7b-12-2024} & 4 & 16 & 2380 & 0 & 52 & 2348 \\
\textsc{Ministral-8B-Instruct-2410} & 0 & 22 & 2378 & 0 & 70 & 2330 \\
\textsc{Mistral-Nemo-Instruct-2407} & 0 & 28 & 2372 & 0 & 1061 & 1339 \\
\textsc{GLM-4-9B-0414} & 0 & 9 & 2391 & 2 & 773 & 1625 \\
\textsc{Qwen3-4B} & 9 & 75 & 2316 & 0 & 203 & 2197 \\
\textsc{Yi-1.5-9B-32K} & 3 & 32 & 2365 & 0 & 1316 & 1084 \\

\hline
\end{tabular}
\caption{Summary of outcomes from all conflicting multilingual haystack retrievals. We report the number of times that the model correctly identifies \textbf{both} answers, or retrieval fails so that \textbf{none} of the answers is identified, or reports only \textbf{one} of the two possible answers without mentioning the other.}
\label{tab:outcomes}
\end{table*}

\subsection{RQ2: presence of language bias}

Given that our results for RQ1 indicate that models overwhelmingly ignore contradicting information and present a single answer, we now turn to RQ2 which asks whether or not the choice of which information to report depends on the language it is presented in. We begin to answer this by looking at how often, over all prompting languages and contrastive haystack pairs, information in one language is chosen over another (full results are found in Appendix~\ref{sec:summary}, with additional haystack sizes in Appendix~\ref{app:pairwise}). Since we work with contrastive pairs of haystacks where all other parameters apart from the languages of the needles (and their immediate context) are equal, we assume that without systematic impact of the language order we would see a binomial distribution for a given pair of languages.
   
For most of the models we see that information in Russian is ignored significantly above chance level (column \textit{rus} of each sub-table). Apart from this, several models display a preference for Chinese (row \textit{cmn} for \textsc{gemma-3-27B}, \textsc{Llama-3.1-8B}, \textsc{gpt-5-nano}, \textsc{Ministral-8B}, \textsc{GLM-4-9B}, \textsc{Yi-1.5-9B}). There are also considerable differences in the overall level of language bias, which is most clearly seen in the sums in the rightmost column, where \textsc{c4ai-command-r7b} has an almost flat distribution over languages whereas for instance \textsc{gpt-5-mini} has 221 instances of English being picked over other languages, compared to just 38 for Russian.

Note that there are a total of 240 contrastive pairs times 5 prompt languages, which makes the maximum sum for a language pair $240 \times 5 = 1200$. We observe that in all cases, counts are considerably lower than that, with a maximum of $93 + 4 = 97$ for English/Russian with the \textsc{gpt-5-mini} model, and only somewhat higher in some of the reduced-haystack complementary experiments reported in Appendix~\ref{app:pairwise}. This implies that factors other than language seem to determine which needle is chosen in at least 90\% of cases. Previous research indicates that these factors include for example the form of (novel) names, since models often prefer the form of certain words over others when answering questions on the basis of conflicting information \citep{kurfali-2025-conflicting}. The significance of our results here is that in the subset of cases where language is the decisive factor, there are clear biases towards certain languages.

While we are mainly interested in the effect of the language that \emph{information} is presented in, we also need to take the language used for \emph{prompting} into account.
For several of the models we see a very clear bias towards preferring information in the prompting language. This is particularly clear in models like {\sc c4ai-command-r7b-12-2024}, which has a low overall level of language bias when averaged over prompt languages, but a substantial bias towards the prompting language. These results show that language biases are even more severe when conditioned on the prompting language. For instance, of the cases when \textsc{gpt-5-nano-2025-08-07} chooses the Russian needle over other languages, this is nearly always when prompting in Russian.
A full table is available as \Fref{tab:prompt25k} in Appendix~\ref{sec:summary}.

\subsection{RQ3: language bias across models}

To test whether some languages are favored over the other languages of our sample, in a way that is consistent across different LLMs, we fit a hierarchical statistical model using Stan \citep{stan}.
\begin{align*}
    s &\sim  \textrm{half-Normal}(0, 100) \\
    b_{l_1,l_2} &\sim  \textrm{Normal}(0, 100) \\
    b_{l_1,l_2,l_p,m} &\sim \textrm{Normal}(b_{l_1,l_2}, s) \\
    w_{l_1,l_2,l_p,m} &\sim \textrm{Bernoulli}(\sigma(b_{l_1,l_2,l_p,m}))
\end{align*}
where $l_1$ and $l_2$ are two different languages that make up a subset of contrastive pairs of bilingual haystacks, $l_p$ is the language used for prompting the model $m$, $w_{l_1,l_2,l_p,m}$ is the binary outcome (win for $l_1$ or for $l_2$) for a given contrastive pair when model $m$ is prompted with language $l_p$. Our main interest is in the parameter $b_{l_1,l_2}$. This represents the log-odds overall bias across models and prompting languages for choosing information from $l_1$ over $l_2$. We have chosen to treat models as independent, even though some of them are from the same family (we have two Llama models and three GPT models). We feel that this is justified, given that we have a relatively small sample of model and that previous work has found that models from the same family can behave very differently in this task \citep{kurfali-2025-conflicting}.

We fit one model for each haystack size. \Fref{tab:biasparam25k} summarizes the posterior of $b_{l_1,l_2}$ for large (up to 25~000 words) haystacks, and \Fref{tab:biasparam1k} for small (up to 1000 words) haystacks. Tables for intermediate sizes can be found in Appendix~\ref{app:biasparam}. In our sample of languages, we see that information in Russian is strongly dispreferred in comparison to all of the remaining languages, and that this applies to all haystack sizes. For the longest haystacks, which is the most challenging version of the task, there is also strong evidence that Chinese is preferred over the other languages, with the possible exception of Turkish.

\begin{table}[t]
\begin{tabular}{lrl}
Language pair $l_1, l_2$ & $P(> 0)$ & 95\% CI \\
\hline
Chinese vs German & 99.0\% & $[0.11, 1.52]$ \\
Chinese vs English & 96.6\% & $[-0.04, 1.28]$ \\
Chinese vs Russian & 100.0\% & $[2.54, 4.17]$ \\
Chinese vs Turkish & 88.4\% & $[-0.27, 1.16]$ \\
German vs English & 58.7\% & $[-0.62, 0.79]$ \\
German vs Russian & 100.0\% & $[1.74, 3.36]$ \\
Turkish vs German & 64.1\% & $[-0.60, 0.88]$ \\
English vs Russian & 100.0\% & $[2.01, 3.64]$ \\
Turkish vs English & 77.3\% & $[-0.44, 0.97]$ \\
Turkish vs Russian & 100.0\% & $[2.39, 4.12]$ \\
\end{tabular}
\caption{Estimates of the log-odds bias parameter $b_{l_1, l_2}$, representing the overall bias (over all prompting languages and models) of language $l_1$ over language $l_2$. Haystack size is 25~000. CI = Bayesian credibility interval.}
\label{tab:biasparam25k}
\end{table}

\begin{table}[t]
\begin{tabular}{lrl}
Language pair $l_1, l_2$ & $P(> 0)$ & 95\% CI \\
\hline
Chinese vs German & 87.5\% & $[-0.27, 1.05]$ \\
English vs Chinese & 52.8\% & $[-0.61, 0.66]$ \\
Chinese vs Russian & 100.0\% & $[2.53, 3.96]$ \\
Chinese vs Turkish & 93.7\% & $[-0.14, 1.16]$ \\
English vs German & 100.0\% & $[0.91, 2.33]$ \\
German vs Russian & 100.0\% & $[2.43, 3.81]$ \\
German vs Turkish & 96.2\% & $[-0.07, 1.24]$ \\
English vs Russian & 100.0\% & $[2.68, 4.16]$ \\
English vs Turkish & 100.0\% & $[0.67, 1.98]$ \\
Turkish vs Russian & 100.0\% & $[2.33, 3.72]$ \\
\end{tabular}
\caption{Estimates of the log-odds bias parameter $b_{l_1, l_2}$, representing the overall bias (over all prompting languages and models) of language $l_1$ over language $l_2$. Haystack size is 1000. CI = Bayesian credibility interval.}
\label{tab:biasparam1k}
\end{table}

\subsection{RQ4: language bias across geographic areas}

Finally, we ask whether the origin of each LLM has an effect on its language bias. To answer this, we extend the model to include a variable for the origin $o_m$ of a particular model, and add the origin $o$ as an additional index to the bias parameter $b_{o,l_1,l_2}$.
\begin{align*}
    s &\sim \textrm{half-Normal}(0, 100) \\
    b_{o,l_1,l_2} &\sim  \textrm{Normal}(0, 100) \\
    b_{l_1,l_2,l_p,m} &\sim \textrm{Normal}(b_{o_m,l_1,l_2}, s) \\
    w_{l_1,l_2,l_p,m} &\sim \textrm{Bernoulli}(\sigma(b_{l_1,l_2,l_p,m}))
\end{align*}

For ease of analysis and due to the limited number of models from each origin, we make a coarse classification only consider two origins: West (USA, Canada, France) and East (PRC).

The language biases are shown separately in \Fref{tab:origineastern} (eastern models) and \Fref{tab:originwestern} (western models). In addition, the difference between these tables are shows in \Fref{tab:origin}. Overall, the general patterns are similar, with a bias against Russian for models from both origins, and a tendency of a bias towards Chinese. In both cases, the pattern is stronger for the eastern models than for the western. One difference to note is that the eastern models have a weak overall bias against English, that is not found in the western models. There is not a single case of Russian being favored over Turkish in any of the eastern models, and combined with our weak prior this leads to a very high estimated bias of Turkish over Russian.

Altogether, there is some evidence of models favoring the dominant language (Chinese and English, respectively) of their origin. On the other hand, there also seems to be a common (in our sample) bias towards Chinese, and a common bias against Russian.

\begin{table}[t]
\begin{tabular}{lrrrrr}
 & cmn & deu & eng & rus & tur \\
\hline
cmn &  & \textcolor{gray}{0.9} & 1.1 & \textbf{2.6} & \textcolor{gray}{0.7} \\
deu & \textcolor{gray}{-0.9} &  & 1.3 & 2.0 & \textcolor{gray}{0.1} \\
eng & -1.1 & -1.3 &  & \textcolor{gray}{0.5} & -1.5 \\
rus & \textbf{-2.6} & -2.0 & \textcolor{gray}{-0.5} &  & \textbf{-77.0} \\
tur & \textcolor{gray}{-0.7} & \textcolor{gray}{-0.1} & 1.5 & \textbf{77.0} &  \\
\hline
\end{tabular}
\caption{Difference in bias among eastern models over western models for choosing $l_1$ (row) over $l_2$ (column).
A \emph{row} with high values indicates a preference for the language of
that row by eastern models, while a \emph{column} with high values indicates
a preference of the column language among western models. Text weight indicates model confidence in the sign of the difference: \textbf{Bold}: $p > 0.99$, black: $p > 0.90$, \textcolor{gray}{gray}: $p \leq 0.90$. Values presented are posterior medians.
Haystack size is 25~000.}
\label{tab:origin}
\end{table}

\begin{table}[t]
\begin{tabular}{lrrrrr}
 & cmn & deu & eng & rus & tur \\
\hline
cmn &  & 1.5 & 1.4 & \textbf{5.5} & 1.0 \\
deu & -1.5 &  & 1.1 & \textbf{4.3} & \textcolor{gray}{-0.0} \\
eng & -1.4 & -1.1 &  & \textbf{3.2} & -1.4 \\
rus & \textbf{-5.5} & \textbf{-4.3} & \textbf{-3.2} &  & \textbf{-79.9} \\
tur & \textcolor{gray}{-1.0} & \textcolor{gray}{0.0} & 1.4 & \textbf{79.9} &  \\
\hline
\end{tabular}
\caption{Bias among eastern models for choosing $l_1$ (row) over $l_2$ (column).
All other details are identical to \Fref{tab:origin}.}
\label{tab:origineastern}
\end{table}

\begin{table}[t]
\begin{tabular}{lrrrrr}
 & cmn & deu & eng & rus & tur \\
\hline
cmn &  & 0.6 & \textcolor{gray}{0.3} & \textbf{2.9} & \textcolor{gray}{0.3} \\
deu & -0.6 &  & \textcolor{gray}{-0.3} & \textbf{2.2} & \textcolor{gray}{-0.2} \\
eng & \textcolor{gray}{-0.3} & \textcolor{gray}{0.3} &  & \textbf{2.7} & \textcolor{gray}{0.1} \\
rus & \textbf{-2.9} & \textbf{-2.2} & \textbf{-2.7} &  & \textbf{-2.8} \\
tur & \textcolor{gray}{-0.3} & \textcolor{gray}{0.2} & \textcolor{gray}{-0.1} & \textbf{2.8} &  \\
\hline
\end{tabular}
\caption{Bias among western models for choosing $l_1$ (row) over $l_2$ (column).
All other details identical to \Fref{tab:origin}.}
\label{tab:originwestern}
\end{table}

\section{Discussion and conclusions}

We found that all LLMs under study were surprisingly consistent in their behavior of asserting a single answer in the presence of conflicting information (RQ1). Although there are some differences between conditions, the same general behavior is observed for both monolingual and multilingual haystacks, across all five languages and 12 models, with different prompts and widely different context lengths.  This is a surprising and worrying mode of failure, in a task of great practical importance that intuitively would seem trivial compared to many of the more complex tasks that LLMs excel at. It falls beyond the scope of this study to identify the mechanisms behind this behavior, but we identify this as an important area of future work. 

We proceeded to ask whether the language that a piece of information is presented in influences the chances of that information to make it into the final LLM answer (RQ2). Based on pairs of haystack configurations that differ only in the languages that each piece of information is presented in, we saw that while language was typically not the decisive factor when selecting among conflicting information, in the cases where this happened there is in many cases a strong tendency for models to prefer one language over another. Furthermore, we found that the language preferences are to a large extent consistent across LLMs (RQ3), and even for LLMs trained in different politico-cultural spheres (RQ4). The strongest pattern is the nearly universal dispreference for information presented in Russian, even though there is no corresponding failure of retrieval in \emph{monolingual} Russian haystacks, which means that results can not simply be explained by models being worse at processing information in Russian. Given our relatively small sample of languages (5), we are unable to tell whether this is unique to Russian or applies to a larger category of languages. One might hypothesize that orthography contributes, since Russian is written in Cyrillic script unlike the majority of Latin-orthography languages, but this is contradicted by the fact that the other universal pattern is a \emph{preference for} the other non-Latin orthography in our data, Chinese. Similarly, our results can hardly be explained by the size of training data alone, when Turkish is often preferred over English. This raises important questions that we leave to future work: what is the pattern of language preferences at a larger scale, and what are the mechanisms that cause this bias? In order to fully answer these questions, one would need a larger sample of languages and of models trained in different parts of the world, as well as the application of methods for investigating the internal representations of models \citep[e.g.,][]{frasertaliente2026nla}.

Our findings mostly agree with those of \citet{wenyi2025chinesechineselanguagemodels}, who found no major differences in the multilingual abilities of PRC-trained LLMs, and those trained elsewhere. This stands in contrast to other work \citep{buyl2026large} which found considerable differences with respect to politically important content. We hypothesize that this difference is mainly due to a general neglect during model training of question answering under conflicting information, especially in multilingual settings. Another important direction of future work is to investigate how changes in model training can either strengthen or weaken the types of language bias we have identified, as well as pinpointing the mechanisms that contribute to unbalanced weighing of information.

\section*{Acknowledgments}

This research is partially funded by the Swedish Research Council through grant agreement no. 2024-01506, and by the Swedish national research infrastructure Språkbanken, jointly financially supported by the Swedish Research Council (2018–2028; grants 2017-00626 and 2023-00161) and the 10 participating partner institutions. The experiments with the open-source LLMs were partially enabled by the National Academic Infrastructure for Supercomputing in Sweden (NAISS) (project: 2025/23-566).

\section*{Limitations}

All articles were originally in English, and predominantly from English-speaking areas. The first names are all typical of males from 1940s Liverpool, and the surnames are chosen to be English-sounding. A broader selection of both needle and haystack data would have allowed investigating the interaction between the topics discussed and the language used for the discussion.

The sample of languages and models are both relatively small, compared to the total number of languages and LLMs. This makes it difficult or impossible to generalize findings beyond our particular sample of languages. In addition, the samples are not as diverse as one would like, with three out of five languages being Indo-European and the majority of LLMs in the 3--9B parameter range.

For comparability with general LLMs and because of the limited compute budget typical for information retrieval tasks, our experiments set `reasoning' (chain-of-thought, CoT) models to use a minimal amount of CoT effort. However, \citet{fu2026reasoningcapabilitysafetylongcontext} found in a broader study on LLM safety that increasing CoT tokens was a strong predictor of identifying harmful intent. It is therefore important to note that our findings may not hold for state-of-the-art models with maximum CoT effort. If a large compute budget is available and focus is set on identifying contradictions, the results by \citet{lovering2025findinginconsistenciesdocuments} indicate that state-of-the-art models can have a reasonably high rate (somewhat over 50\%) of identifying conflicting information.

Although we investigate the effect of prompting language as part of our answer to RQ2, we abstain from performing the rest of the analyses in this paper separately for the five different prompting languages. By instead averaging over all prompting languages, we obtain a somewhat conservative estimate of the amount of language bias, since biases due to the prompting language are averaged out.

We note that the results of our research could potentially be applied to harmful activities such as promoting disinformation. By careful selection of which language the disinformation is presented in, and considering the languages likely used for conflicting (factual) information and LLM prompting, it is possible to increase the likelihood of disinformation being chosen in LLM-generated summaries or answers. Yet, we believe that the advantages of publishing this research outweigh the disadvantages, because only by bringing attention to the problem can we spur research on effective countermeasures.

\bibliography{custom}

@misc{stan,
title={Stan Reference Manual, 2.38.0},
author={Stan Development Team, The},
year={2026},
url={https://mc-stan.org},
}

@misc{gemmateam2025gemma3technicalreport,
      title={Gemma 3 Technical Report}, 
      author={GemmaTeam and Aishwarya Kamath and Johan Ferret and Shreya Pathak and Nino Vieillard and Ramona Merhej and Sarah Perrin and Tatiana Matejovicova and Alexandre Ramé and Morgane Rivière and Louis Rouillard and Thomas Mesnard and Geoffrey Cideron and Jean-bastien Grill and Sabela Ramos and Edouard Yvinec and Michelle Casbon and Etienne Pot and Ivo Penchev and Gaël Liu and Francesco Visin and Kathleen Kenealy and Lucas Beyer and Xiaohai Zhai and Anton Tsitsulin and Robert Busa-Fekete and Alex Feng and Noveen Sachdeva and Benjamin Coleman and Yi Gao and Basil Mustafa and Iain Barr and Emilio Parisotto and David Tian and Matan Eyal and Colin Cherry and Jan-Thorsten Peter and Danila Sinopalnikov and Surya Bhupatiraju and Rishabh Agarwal and Mehran Kazemi and Dan Malkin and Ravin Kumar and David Vilar and Idan Brusilovsky and Jiaming Luo and Andreas Steiner and Abe Friesen and Abhanshu Sharma and Abheesht Sharma and Adi Mayrav Gilady and Adrian Goedeckemeyer and Alaa Saade and Alex Feng and Alexander Kolesnikov and Alexei Bendebury and Alvin Abdagic and Amit Vadi and András György and André Susano Pinto and Anil Das and Ankur Bapna and Antoine Miech and Antoine Yang and Antonia Paterson and Ashish Shenoy and Ayan Chakrabarti and Bilal Piot and Bo Wu and Bobak Shahriari and Bryce Petrini and Charlie Chen and Charline Le Lan and Christopher A. Choquette-Choo and CJ Carey and Cormac Brick and Daniel Deutsch and Danielle Eisenbud and Dee Cattle and Derek Cheng and Dimitris Paparas and Divyashree Shivakumar Sreepathihalli and Doug Reid and Dustin Tran and Dustin Zelle and Eric Noland and Erwin Huizenga and Eugene Kharitonov and Frederick Liu and Gagik Amirkhanyan and Glenn Cameron and Hadi Hashemi and Hanna Klimczak-Plucińska and Harman Singh and Harsh Mehta and Harshal Tushar Lehri and Hussein Hazimeh and Ian Ballantyne and Idan Szpektor and Ivan Nardini and Jean Pouget-Abadie and Jetha Chan and Joe Stanton and John Wieting and Jonathan Lai and Jordi Orbay and Joseph Fernandez and Josh Newlan and Ju-yeong Ji and Jyotinder Singh and Kat Black and Kathy Yu and Kevin Hui and Kiran Vodrahalli and Klaus Greff and Linhai Qiu and Marcella Valentine and Marina Coelho and Marvin Ritter and Matt Hoffman and Matthew Watson and Mayank Chaturvedi and Michael Moynihan and Min Ma and Nabila Babar and Natasha Noy and Nathan Byrd and Nick Roy and Nikola Momchev and Nilay Chauhan and Noveen Sachdeva and Oskar Bunyan and Pankil Botarda and Paul Caron and Paul Kishan Rubenstein and Phil Culliton and Philipp Schmid and Pier Giuseppe Sessa and Pingmei Xu and Piotr Stanczyk and Pouya Tafti and Rakesh Shivanna and Renjie Wu and Renke Pan and Reza Rokni and Rob Willoughby and Rohith Vallu and Ryan Mullins and Sammy Jerome and Sara Smoot and Sertan Girgin and Shariq Iqbal and Shashir Reddy and Shruti Sheth and Siim Põder and Sijal Bhatnagar and Sindhu Raghuram Panyam and Sivan Eiger and Susan Zhang and Tianqi Liu and Trevor Yacovone and Tyler Liechty and Uday Kalra and Utku Evci and Vedant Misra and Vincent Roseberry and Vlad Feinberg and Vlad Kolesnikov and Woohyun Han and Woosuk Kwon and Xi Chen and Yinlam Chow and Yuvein Zhu and Zichuan Wei and Zoltan Egyed and Victor Cotruta and Minh Giang and Phoebe Kirk and Anand Rao and Kat Black and Nabila Babar and Jessica Lo and Erica Moreira and Luiz Gustavo Martins and Omar Sanseviero and Lucas Gonzalez and Zach Gleicher and Tris Warkentin and Vahab Mirrokni and Evan Senter and Eli Collins and Joelle Barral and Zoubin Ghahramani and Raia Hadsell and Yossi Matias and D. Sculley and Slav Petrov and Noah Fiedel and Noam Shazeer and Oriol Vinyals and Jeff Dean and Demis Hassabis and Koray Kavukcuoglu and Clement Farabet and Elena Buchatskaya and Jean-Baptiste Alayrac and Rohan Anil and Dmitry and Lepikhin and Sebastian Borgeaud and Olivier Bachem and Armand Joulin and Alek Andreev and Cassidy Hardin and Robert Dadashi and Léonard Hussenot},
      year={2025},
      eprint={2503.19786},
      archivePrefix={arXiv},
      primaryClass={cs.CL},
      url={https://arxiv.org/abs/2503.19786}, 
}

@misc{ai2025yiopenfoundationmodels,
      title={Yi: Open Foundation Models by 01.AI}, 
      author={01AI and Alex Young and Bei Chen and Chao Li and Chengen Huang and Ge Zhang and Guanwei Zhang and Guoyin Wang and Heng Li and Jiangcheng Zhu and Jianqun Chen and Jing Chang and Kaidong Yu and Peng Liu and Qiang Liu and Shawn Yue and Senbin Yang and Shiming Yang and Wen Xie and Wenhao Huang and Xiaohui Hu and Xiaoyi Ren and Xinyao Niu and Pengcheng Nie and Yanpeng Li and Yuchi Xu and Yudong Liu and Yue Wang and Yuxuan Cai and Zhenyu Gu and Zhiyuan Liu and Zonghong Dai},
      year={2025},
      eprint={2403.04652},
      archivePrefix={arXiv},
      primaryClass={cs.CL},
      url={https://arxiv.org/abs/2403.04652}, 
}

@misc{yang2025qwen3technicalreport,
      title={Qwen3 Technical Report}, 
      author={An Yang and Anfeng Li and Baosong Yang and Beichen Zhang and Binyuan Hui and Bo Zheng and Bowen Yu and Chang Gao and Chengen Huang and Chenxu Lv and Chujie Zheng and Dayiheng Liu and Fan Zhou and Fei Huang and Feng Hu and Hao Ge and Haoran Wei and Huan Lin and Jialong Tang and Jian Yang and Jianhong Tu and Jianwei Zhang and Jianxin Yang and Jiaxi Yang and Jing Zhou and Jingren Zhou and Junyang Lin and Kai Dang and Keqin Bao and Kexin Yang and Le Yu and Lianghao Deng and Mei Li and Mingfeng Xue and Mingze Li and Pei Zhang and Peng Wang and Qin Zhu and Rui Men and Ruize Gao and Shixuan Liu and Shuang Luo and Tianhao Li and Tianyi Tang and Wenbiao Yin and Xingzhang Ren and Xinyu Wang and Xinyu Zhang and Xuancheng Ren and Yang Fan and Yang Su and Yichang Zhang and Yinger Zhang and Yu Wan and Yuqiong Liu and Zekun Wang and Zeyu Cui and Zhenru Zhang and Zhipeng Zhou and Zihan Qiu},
      year={2025},
      eprint={2505.09388},
      archivePrefix={arXiv},
      primaryClass={cs.CL},
      url={https://arxiv.org/abs/2505.09388}, 
}

@misc{grattafiori2024llama3herdmodels,
      title={The Llama 3 Herd of Models}, 
      author={Aaron Grattafiori and Abhimanyu Dubey and Abhinav Jauhri and Abhinav Pandey and Abhishek Kadian and Ahmad Al-Dahle and Aiesha Letman and Akhil Mathur and Alan Schelten and Alex Vaughan and Amy Yang and Angela Fan and Anirudh Goyal and Anthony Hartshorn and Aobo Yang and Archi Mitra and Archie Sravankumar and Artem Korenev and Arthur Hinsvark and Arun Rao and Aston Zhang and Aurelien Rodriguez and Austen Gregerson and Ava Spataru and Baptiste Roziere and Bethany Biron and Binh Tang and Bobbie Chern and Charlotte Caucheteux and Chaya Nayak and Chloe Bi and Chris Marra and Chris McConnell and Christian Keller and Christophe Touret and Chunyang Wu and Corinne Wong and Cristian Canton Ferrer and Cyrus Nikolaidis and Damien Allonsius and Daniel Song and Danielle Pintz and Danny Livshits and Danny Wyatt and David Esiobu and Dhruv Choudhary and Dhruv Mahajan and Diego Garcia-Olano and Diego Perino and Dieuwke Hupkes and Egor Lakomkin and Ehab AlBadawy and Elina Lobanova and Emily Dinan and Eric Michael Smith and Filip Radenovic and Francisco Guzmán and Frank Zhang and Gabriel Synnaeve and Gabrielle Lee and Georgia Lewis Anderson and Govind Thattai and Graeme Nail and Gregoire Mialon and Guan Pang and Guillem Cucurell and Hailey Nguyen and Hannah Korevaar and Hu Xu and Hugo Touvron and Iliyan Zarov and Imanol Arrieta Ibarra and Isabel Kloumann and Ishan Misra and Ivan Evtimov and Jack Zhang and Jade Copet and Jaewon Lee and Jan Geffert and Jana Vranes and Jason Park and Jay Mahadeokar and Jeet Shah and Jelmer van der Linde and Jennifer Billock and Jenny Hong and Jenya Lee and Jeremy Fu and Jianfeng Chi and Jianyu Huang and Jiawen Liu and Jie Wang and Jiecao Yu and Joanna Bitton and Joe Spisak and Jongsoo Park and Joseph Rocca and Joshua Johnstun and Joshua Saxe and Junteng Jia and Kalyan Vasuden Alwala and Karthik Prasad and Kartikeya Upasani and Kate Plawiak and Ke Li and Kenneth Heafield and Kevin Stone and Khalid El-Arini and Krithika Iyer and Kshitiz Malik and Kuenley Chiu and Kunal Bhalla and Kushal Lakhotia and Lauren Rantala-Yeary and Laurens van der Maaten and Lawrence Chen and Liang Tan and Liz Jenkins and Louis Martin and Lovish Madaan and Lubo Malo and Lukas Blecher and Lukas Landzaat and Luke de Oliveira and Madeline Muzzi and Mahesh Pasupuleti and Mannat Singh and Manohar Paluri and Marcin Kardas and Maria Tsimpoukelli and Mathew Oldham and Mathieu Rita and Maya Pavlova and Melanie Kambadur and Mike Lewis and Min Si and Mitesh Kumar Singh and Mona Hassan and Naman Goyal and Narjes Torabi and Nikolay Bashlykov and Nikolay Bogoychev and Niladri Chatterji and Ning Zhang and Olivier Duchenne and Onur Çelebi and Patrick Alrassy and Pengchuan Zhang and Pengwei Li and Petar Vasic and Peter Weng and Prajjwal Bhargava and Pratik Dubal and Praveen Krishnan and Punit Singh Koura and Puxin Xu and Qing He and Qingxiao Dong and Ragavan Srinivasan and Raj Ganapathy and Ramon Calderer and Ricardo Silveira Cabral and Robert Stojnic and Roberta Raileanu and Rohan Maheswari and Rohit Girdhar and Rohit Patel and Romain Sauvestre and Ronnie Polidoro and Roshan Sumbaly and Ross Taylor and Ruan Silva and Rui Hou and Rui Wang and Saghar Hosseini and Sahana Chennabasappa and Sanjay Singh and Sean Bell and Seohyun Sonia Kim and Sergey Edunov and Shaoliang Nie and Sharan Narang and Sharath Raparthy and Sheng Shen and Shengye Wan and Shruti Bhosale and Shun Zhang and Simon Vandenhende and Soumya Batra and Spencer Whitman and Sten Sootla and Stephane Collot and Suchin Gururangan and Sydney Borodinsky and Tamar Herman and Tara Fowler and Tarek Sheasha and Thomas Georgiou and Thomas Scialom and Tobias Speckbacher and Todor Mihaylov and Tong Xiao and Ujjwal Karn and Vedanuj Goswami and Vibhor Gupta and Vignesh Ramanathan and Viktor Kerkez and Vincent Gonguet and Virginie Do and Vish Vogeti and Vítor Albiero and Vladan Petrovic and Weiwei Chu and Wenhan Xiong and Wenyin Fu and Whitney Meers and Xavier Martinet and Xiaodong Wang and Xiaofang Wang and Xiaoqing Ellen Tan and Xide Xia and Xinfeng Xie and Xuchao Jia and Xuewei Wang and Yaelle Goldschlag and Yashesh Gaur and Yasmine Babaei and Yi Wen and Yiwen Song and Yuchen Zhang and Yue Li and Yuning Mao and Zacharie Delpierre Coudert and Zheng Yan and Zhengxing Chen and Zoe Papakipos and Aaditya Singh and Aayushi Srivastava and Abha Jain and Adam Kelsey and Adam Shajnfeld and Adithya Gangidi and Adolfo Victoria and Ahuva Goldstand and Ajay Menon and Ajay Sharma and Alex Boesenberg and Alexei Baevski and Allie Feinstein and Amanda Kallet and Amit Sangani and Amos Teo and Anam Yunus and Andrei Lupu and Andres Alvarado and Andrew Caples and Andrew Gu and Andrew Ho and Andrew Poulton and Andrew Ryan and Ankit Ramchandani and Annie Dong and Annie Franco and Anuj Goyal and Aparajita Saraf and Arkabandhu Chowdhury and Ashley Gabriel and Ashwin Bharambe and Assaf Eisenman and Azadeh Yazdan and Beau James and Ben Maurer and Benjamin Leonhardi and Bernie Huang and Beth Loyd and Beto De Paola and Bhargavi Paranjape and Bing Liu and Bo Wu and Boyu Ni and Braden Hancock and Bram Wasti and Brandon Spence and Brani Stojkovic and Brian Gamido and Britt Montalvo and Carl Parker and Carly Burton and Catalina Mejia and Ce Liu and Changhan Wang and Changkyu Kim and Chao Zhou and Chester Hu and Ching-Hsiang Chu and Chris Cai and Chris Tindal and Christoph Feichtenhofer and Cynthia Gao and Damon Civin and Dana Beaty and Daniel Kreymer and Daniel Li and David Adkins and David Xu and Davide Testuggine and Delia David and Devi Parikh and Diana Liskovich and Didem Foss and Dingkang Wang and Duc Le and Dustin Holland and Edward Dowling and Eissa Jamil and Elaine Montgomery and Eleonora Presani and Emily Hahn and Emily Wood and Eric-Tuan Le and Erik Brinkman and Esteban Arcaute and Evan Dunbar and Evan Smothers and Fei Sun and Felix Kreuk and Feng Tian and Filippos Kokkinos and Firat Ozgenel and Francesco Caggioni and Frank Kanayet and Frank Seide and Gabriela Medina Florez and Gabriella Schwarz and Gada Badeer and Georgia Swee and Gil Halpern and Grant Herman and Grigory Sizov and Guangyi and Zhang and Guna Lakshminarayanan and Hakan Inan and Hamid Shojanazeri and Han Zou and Hannah Wang and Hanwen Zha and Haroun Habeeb and Harrison Rudolph and Helen Suk and Henry Aspegren and Hunter Goldman and Hongyuan Zhan and Ibrahim Damlaj and Igor Molybog and Igor Tufanov and Ilias Leontiadis and Irina-Elena Veliche and Itai Gat and Jake Weissman and James Geboski and James Kohli and Janice Lam and Japhet Asher and Jean-Baptiste Gaya and Jeff Marcus and Jeff Tang and Jennifer Chan and Jenny Zhen and Jeremy Reizenstein and Jeremy Teboul and Jessica Zhong and Jian Jin and Jingyi Yang and Joe Cummings and Jon Carvill and Jon Shepard and Jonathan McPhie and Jonathan Torres and Josh Ginsburg and Junjie Wang and Kai Wu and Kam Hou U and Karan Saxena and Kartikay Khandelwal and Katayoun Zand and Kathy Matosich and Kaushik Veeraraghavan and Kelly Michelena and Keqian Li and Kiran Jagadeesh and Kun Huang and Kunal Chawla and Kyle Huang and Lailin Chen and Lakshya Garg and Lavender A and Leandro Silva and Lee Bell and Lei Zhang and Liangpeng Guo and Licheng Yu and Liron Moshkovich and Luca Wehrstedt and Madian Khabsa and Manav Avalani and Manish Bhatt and Martynas Mankus and Matan Hasson and Matthew Lennie and Matthias Reso and Maxim Groshev and Maxim Naumov and Maya Lathi and Meghan Keneally and Miao Liu and Michael L. Seltzer and Michal Valko and Michelle Restrepo and Mihir Patel and Mik Vyatskov and Mikayel Samvelyan and Mike Clark and Mike Macey and Mike Wang and Miquel Jubert Hermoso and Mo Metanat and Mohammad Rastegari and Munish Bansal and Nandhini Santhanam and Natascha Parks and Natasha White and Navyata Bawa and Nayan Singhal and Nick Egebo and Nicolas Usunier and Nikhil Mehta and Nikolay Pavlovich Laptev and Ning Dong and Norman Cheng and Oleg Chernoguz and Olivia Hart and Omkar Salpekar and Ozlem Kalinli and Parkin Kent and Parth Parekh and Paul Saab and Pavan Balaji and Pedro Rittner and Philip Bontrager and Pierre Roux and Piotr Dollar and Polina Zvyagina and Prashant Ratanchandani and Pritish Yuvraj and Qian Liang and Rachad Alao and Rachel Rodriguez and Rafi Ayub and Raghotham Murthy and Raghu Nayani and Rahul Mitra and Rangaprabhu Parthasarathy and Raymond Li and Rebekkah Hogan and Robin Battey and Rocky Wang and Russ Howes and Ruty Rinott and Sachin Mehta and Sachin Siby and Sai Jayesh Bondu and Samyak Datta and Sara Chugh and Sara Hunt and Sargun Dhillon and Sasha Sidorov and Satadru Pan and Saurabh Mahajan and Saurabh Verma and Seiji Yamamoto and Sharadh Ramaswamy and Shaun Lindsay and Shaun Lindsay and Sheng Feng and Shenghao Lin and Shengxin Cindy Zha and Shishir Patil and Shiva Shankar and Shuqiang Zhang and Shuqiang Zhang and Sinong Wang and Sneha Agarwal and Soji Sajuyigbe and Soumith Chintala and Stephanie Max and Stephen Chen and Steve Kehoe and Steve Satterfield and Sudarshan Govindaprasad and Sumit Gupta and Summer Deng and Sungmin Cho and Sunny Virk and Suraj Subramanian and Sy Choudhury and Sydney Goldman and Tal Remez and Tamar Glaser and Tamara Best and Thilo Koehler and Thomas Robinson and Tianhe Li and Tianjun Zhang and Tim Matthews and Timothy Chou and Tzook Shaked and Varun Vontimitta and Victoria Ajayi and Victoria Montanez and Vijai Mohan and Vinay Satish Kumar and Vishal Mangla and Vlad Ionescu and Vlad Poenaru and Vlad Tiberiu Mihailescu and Vladimir Ivanov and Wei Li and Wenchen Wang and Wenwen Jiang and Wes Bouaziz and Will Constable and Xiaocheng Tang and Xiaojian Wu and Xiaolan Wang and Xilun Wu and Xinbo Gao and Yaniv Kleinman and Yanjun Chen and Ye Hu and Ye Jia and Ye Qi and Yenda Li and Yilin Zhang and Ying Zhang and Yossi Adi and Youngjin Nam and Yu and Wang and Yu Zhao and Yuchen Hao and Yundi Qian and Yunlu Li and Yuzi He and Zach Rait and Zachary DeVito and Zef Rosnbrick and Zhaoduo Wen and Zhenyu Yang and Zhiwei Zhao and Zhiyu Ma},
      year={2024},
      eprint={2407.21783},
      archivePrefix={arXiv},
      primaryClass={cs.AI},
      url={https://arxiv.org/abs/2407.21783}, 
}

@misc{glm2024chatglm,
      title={ChatGLM: A Family of Large Language Models from GLM-130B to GLM-4 All Tools},
      author={Team GLM and Aohan Zeng and Bin Xu and Bowen Wang and Chenhui Zhang and Da Yin and Diego Rojas and Guanyu Feng and Hanlin Zhao and Hanyu Lai and Hao Yu and Hongning Wang and Jiadai Sun and Jiajie Zhang and Jiale Cheng and Jiayi Gui and Jie Tang and Jing Zhang and Juanzi Li and Lei Zhao and Lindong Wu and Lucen Zhong and Mingdao Liu and Minlie Huang and Peng Zhang and Qinkai Zheng and Rui Lu and Shuaiqi Duan and Shudan Zhang and Shulin Cao and Shuxun Yang and Weng Lam Tam and Wenyi Zhao and Xiao Liu and Xiao Xia and Xiaohan Zhang and Xiaotao Gu and Xin Lv and Xinghan Liu and Xinyi Liu and Xinyue Yang and Xixuan Song and Xunkai Zhang and Yifan An and Yifan Xu and Yilin Niu and Yuantao Yang and Yueyan Li and Yushi Bai and Yuxiao Dong and Zehan Qi and Zhaoyu Wang and Zhen Yang and Zhengxiao Du and Zhenyu Hou and Zihan Wang},
      year={2024},
      eprint={2406.12793},
      archivePrefix={arXiv},
      primaryClass={id='cs.CL' full_name='Computation and Language' is_active=True alt_name='cmp-lg' in_archive='cs' is_general=False description='Covers natural language processing. Roughly includes material in ACM Subject Class I.2.7. Note that work on artificial languages (programming languages, logics, formal systems) that does not explicitly address natural-language issues broadly construed (natural-language processing, computational linguistics, speech, text retrieval, etc.) is not appropriate for this area.'}
}

@misc{cohere2025commandaenterprisereadylarge,
      title={Command A: An Enterprise-Ready Large Language Model}, 
      author={Team Cohere and : and Aakanksha and Arash Ahmadian and Marwan Ahmed and Jay Alammar and Milad Alizadeh and Yazeed Alnumay and Sophia Althammer and Arkady Arkhangorodsky and Viraat Aryabumi and Dennis Aumiller and Raphaël Avalos and Zahara Aviv and Sammie Bae and Saurabh Baji and Alexandre Barbet and Max Bartolo and Björn Bebensee and Neeral Beladia and Walter Beller-Morales and Alexandre Bérard and Andrew Berneshawi and Anna Bialas and Phil Blunsom and Matt Bobkin and Adi Bongale and Sam Braun and Maxime Brunet and Samuel Cahyawijaya and David Cairuz and Jon Ander Campos and Cassie Cao and Kris Cao and Roman Castagné and Julián Cendrero and Leila Chan Currie and Yash Chandak and Diane Chang and Giannis Chatziveroglou and Hongyu Chen and Claire Cheng and Alexis Chevalier and Justin T. Chiu and Eugene Cho and Eugene Choi and Eujeong Choi and Tim Chung and Volkan Cirik and Ana Cismaru and Pierre Clavier and Henry Conklin and Lucas Crawhall-Stein and Devon Crouse and Andres Felipe Cruz-Salinas and Ben Cyrus and Daniel D'souza and Hugo Dalla-Torre and John Dang and William Darling and Omar Darwiche Domingues and Saurabh Dash and Antoine Debugne and Théo Dehaze and Shaan Desai and Joan Devassy and Rishit Dholakia and Kyle Duffy and Ali Edalati and Ace Eldeib and Abdullah Elkady and Sarah Elsharkawy and Irem Ergün and Beyza Ermis and Marzieh Fadaee and Boyu Fan and Lucas Fayoux and Yannis Flet-Berliac and Nick Frosst and Matthias Gallé and Wojciech Galuba and Utsav Garg and Matthieu Geist and Mohammad Gheshlaghi Azar and Ellen Gilsenan-McMahon and Seraphina Goldfarb-Tarrant and Tomas Goldsack and Aidan Gomez and Victor Machado Gonzaga and Nithya Govindarajan and Manoj Govindassamy and Nathan Grinsztajn and Nikolas Gritsch and Patrick Gu and Shangmin Guo and Kilian Haefeli and Rod Hajjar and Tim Hawes and Jingyi He and Sebastian Hofstätter and Sungjin Hong and Sara Hooker and Tom Hosking and Stephanie Howe and Eric Hu and Renjie Huang and Hemant Jain and Ritika Jain and Nick Jakobi and Madeline Jenkins and JJ Jordan and Dhruti Joshi and Jason Jung and Trushant Kalyanpur and Siddhartha Rao Kamalakara and Julia Kedrzycki and Gokce Keskin and Edward Kim and Joon Kim and Wei-Yin Ko and Tom Kocmi and Michael Kozakov and Wojciech Kryściński and Arnav Kumar Jain and Komal Kumar Teru and Sander Land and Michael Lasby and Olivia Lasche and Justin Lee and Patrick Lewis and Jeffrey Li and Jonathan Li and Hangyu Lin and Acyr Locatelli and Kevin Luong and Raymond Ma and Lukáš Mach and Marina Machado and Joanne Magbitang and Brenda Malacara Lopez and Aryan Mann and Kelly Marchisio and Olivia Markham and Alexandre Matton and Alex McKinney and Dominic McLoughlin and Jozef Mokry and Adrien Morisot and Autumn Moulder and Harry Moynehan and Maximilian Mozes and Vivek Muppalla and Lidiya Murakhovska and Hemangani Nagarajan and Alekhya Nandula and Hisham Nasir and Shauna Nehra and Josh Netto-Rosen and Daniel Ohashi and James Owers-Bardsley and Jason Ozuzu and Dennis Padilla and Gloria Park and Sam Passaglia and Jeremy Pekmez and Laura Penstone and Aleksandra Piktus and Case Ploeg and Andrew Poulton and Youran Qi and Shubha Raghvendra and Miguel Ramos and Ekagra Ranjan and Pierre Richemond and Cécile Robert-Michon and Aurélien Rodriguez and Sudip Roy and Sebastian Ruder and Laura Ruis and Louise Rust and Anubhav Sachan and Alejandro Salamanca and Kailash Karthik Saravanakumar and Isha Satyakam and Alice Schoenauer Sebag and Priyanka Sen and Sholeh Sepehri and Preethi Seshadri and Ye Shen and Tom Sherborne and Sylvie Shang Shi and Sanal Shivaprasad and Vladyslav Shmyhlo and Anirudh Shrinivason and Inna Shteinbuk and Amir Shukayev and Mathieu Simard and Ella Snyder and Ava Spataru and Victoria Spooner and Trisha Starostina and Florian Strub and Yixuan Su and Jimin Sun and Dwarak Talupuru and Eugene Tarassov and Elena Tommasone and Jennifer Tracey and Billy Trend and Evren Tumer and Ahmet Üstün and Bharat Venkitesh and David Venuto and Pat Verga and Maxime Voisin and Alex Wang and Donglu Wang and Shijian Wang and Edmond Wen and Naomi White and Jesse Willman and Marysia Winkels and Chen Xia and Jessica Xie and Minjie Xu and Bowen Yang and Tan Yi-Chern and Ivan Zhang and Zhenyu Zhao and Zhoujie Zhao},
      year={2025},
      eprint={2504.00698},
      archivePrefix={arXiv},
      primaryClass={cs.CL},
      url={https://arxiv.org/abs/2504.00698}, 
}

@article{frasertaliente2026nla,
  author={Fraser-Taliente, Kit and Kantamneni, Subhash and Ong, Euan and Mossing, Dan and Lu, Christina and Bogdan, Paul C. and Ameisen, Emmanuel and Chen, James and Kishylau, Dzmitry and Pearce, Adam and Tarng, Julius and Wu, Alex and Wu, Jeff and Zhang, Yang and Ziegler, Daniel M. and Hubinger, Evan and Batson, Joshua and Lindsey, Jack and Zimmerman, Samuel and Marks, Samuel},
  title={Natural Language Autoencoders Produce Unsupervised Explanations of LLM Activations},
  journal={Transformer Circuits Thread},
  year={2026},
  url={https://transformer-circuits.pub/2026/nla/index.html}
}

@misc{tan2026improvedevidenceextractiondocument,
      title={Improved Evidence Extraction for Document Inconsistency Detection with LLMs}, 
      author={Nelvin Tan and Yaowen Zhang and James Asikin Cheung and Fusheng Liu and Yu-Ching Shih and Dong Yang},
      year={2026},
      eprint={2601.02627},
      archivePrefix={arXiv},
      primaryClass={cs.CL},
      url={https://arxiv.org/abs/2601.02627}, 
}

@inproceedings{li-etal-2024-contradoc,
    title = "{C}ontra{D}oc: Understanding Self-Contradictions in Documents with Large Language Models",
    author = "Li, Jierui  and
      Raheja, Vipul  and
      Kumar, Dhruv",
    editor = "Duh, Kevin  and
      Gomez, Helena  and
      Bethard, Steven",
    booktitle = "Proceedings of the 2024 Conference of the North American Chapter of the Association for Computational Linguistics: Human Language Technologies (Volume 1: Long Papers)",
    month = jun,
    year = "2024",
    address = "Mexico City, Mexico",
    publisher = "Association for Computational Linguistics",
    url = "https://aclanthology.org/2024.naacl-long.362/",
    doi = "10.18653/v1/2024.naacl-long.362",
    pages = "6509--6523"
}

@misc{kamradt2023needle,
  author       = {Gregory Kamradt},
  title        = {Needle In A Haystack - Pressure Testing LLMs},
  year         = {2023},
  howpublished = {\url{https://github.com/gkamradt/needle-in-a-haystack}},
  note         = {GitHub repository},
}

@misc{fu2026reasoningcapabilitysafetylongcontext,
      title={Is Reasoning Capability Enough for Safety in Long-Context Language Models?}, 
      author={Yu Fu and Haz Sameen Shahgir and Huanli Gong and Zhipeng Wei and N. Benjamin Erichson and Yue Dong},
      year={2026},
      eprint={2602.08874},
      archivePrefix={arXiv},
      primaryClass={cs.CL},
      url={https://arxiv.org/abs/2602.08874}, 
}

@misc{lovering2025findinginconsistenciesdocuments,
      title={On Finding Inconsistencies in Documents}, 
      author={Charles J. Lovering and Seth Ebner and Brandon Smock and Michael Krumdick and Saad Rabbani and Ahmed Muhammad and Varshini Reddy and Chris Tanner},
      year={2025},
      eprint={2512.18601},
      archivePrefix={arXiv},
      primaryClass={cs.CL},
      url={https://arxiv.org/abs/2512.18601}, 
}

@inproceedings{shaier-etal-2024-adaptive,
    title = "Adaptive Question Answering: Enhancing Language Model Proficiency for Addressing Knowledge Conflicts with Source Citations",
    author = "Shaier, Sagi  and
      Kobren, Ari  and
      Ogren, Philip V.",
    editor = "Al-Onaizan, Yaser  and
      Bansal, Mohit  and
      Chen, Yun-Nung",
    booktitle = "Proceedings of the 2024 Conference on Empirical Methods in Natural Language Processing",
    month = nov,
    year = "2024",
    address = "Miami, Florida, USA",
    publisher = "Association for Computational Linguistics",
    url = "https://aclanthology.org/2024.emnlp-main.956/",
    doi = "10.18653/v1/2024.emnlp-main.956",
    pages = "17226--17239"
}

@article{schuster2026factswinllmsource,
      title={Whose Facts Win? LLM Source Preferences under Knowledge Conflicts}, 
      author={Jakob Schuster and Vagrant Gautam and Katja Markert},
      year={2026},
      eprint={2601.03746},
      archivePrefix={arXiv},
      primaryClass={cs.CL},
      url={https://arxiv.org/abs/2601.03746}, 
}

@article{buyl2026large,
  title={Large language models reflect the ideology of their creators},
  author={Buyl, Maarten and Rogiers, Alexander and Noels, Sander and Bied, Guillaume and Dominguez-Catena, Iris and Heiter, Edith and Johary, Iman and Mara, Alexandru-Cristian and Romero, Rapha{\"e}l and Lijffijt, Jefrey and others},
  journal={npj Artificial Intelligence},
  volume={2},
  number={1},
  pages={7},
  year={2026},
  publisher={Nature Publishing Group UK London}
}

@misc{wenyi2025chinesechineselanguagemodels,
      title={How Chinese are Chinese Language Models? The Puzzling Lack of Language Policy in China's LLMs}, 
      author={Andrea W Wen-Yi and Unso Eun Seo Jo and Lu Jia Lin and David Mimno},
      year={2025},
      eprint={2407.09652},
      archivePrefix={arXiv},
      primaryClass={cs.CL},
      url={https://arxiv.org/abs/2407.09652}, 
}

@misc{hsieh2024rulerwhatsrealcontext,
      title={RULER: What's the Real Context Size of Your Long-Context Language Models?}, 
      author={Cheng-Ping Hsieh and Simeng Sun and Samuel Kriman and Shantanu Acharya and Dima Rekesh and Fei Jia and Yang Zhang and Boris Ginsburg},
      year={2024},
      eprint={2404.06654},
      archivePrefix={arXiv},
      primaryClass={cs.CL},
      url={https://arxiv.org/abs/2404.06654}, 
}

@inproceedings{Wang2024multimodalhaystack,
 author = {Wang, Weiyun and Zhang, Shuibo and Ren, Yiming and Duan, Yuchen and Li, Tiantong and Liu, Shuo and Hu, Mengkang and Chen, Zhe and Zhang, Kaipeng and Lu, Lewei and Zhu, Xizhou and Luo, Ping and Qiao, Yu and Dai, Jifeng and Shao, Wenqi and Wang, Wenhai},
 booktitle = {Advances in Neural Information Processing Systems},
 doi = {10.52202/079017-0649},
 editor = {A. Globerson and L. Mackey and D. Belgrave and A. Fan and U. Paquet and J. Tomczak and C. Zhang},
 pages = {20540--20565},
 publisher = {Curran Associates, Inc.},
 title = {Needle In A Multimodal Haystack},
 url = {https://proceedings.neurips.cc/paper_files/paper/2024/file/24a8968affe71ffe4067d022b9d16566-Paper-Datasets_and_Benchmarks_Track.pdf},
 volume = {37},
 year = {2024}
}

@inproceedings{kurfali-2025-conflicting,
    title = "Conflicting Needles in a Haystack: How {LLM}s behave when faced with contradictory information",
    author = {Kurfali, Murathan  and
      {\"O}stling, Robert},
    editor = "Christodoulopoulos, Christos  and
      Chakraborty, Tanmoy  and
      Rose, Carolyn  and
      Peng, Violet",
    booktitle = "Proceedings of the 2025 Conference on Empirical Methods in Natural Language Processing",
    month = nov,
    year = "2025",
    address = "Suzhou, China",
    publisher = "Association for Computational Linguistics",
    url = "https://aclanthology.org/2025.emnlp-main.1742/",
    doi = "10.18653/v1/2025.emnlp-main.1742",
    pages = "34349--34364",
    ISBN = "979-8-89176-332-6"
}

@inproceedings{cui-etal-2025-multilingual,
    title = "Multilingual Machine Translation with Open Large Language Models at Practical Scale: An Empirical Study",
    author = "Cui, Menglong  and
      Gao, Pengzhi  and
      Liu, Wei  and
      Luan, Jian  and
      Wang, Bin",
    editor = "Chiruzzo, Luis  and
      Ritter, Alan  and
      Wang, Lu",
    booktitle = "Proceedings of the 2025 Conference of the Nations of the Americas Chapter of the Association for Computational Linguistics: Human Language Technologies (Volume 1: Long Papers)",
    month = apr,
    year = "2025",
    address = "Albuquerque, New Mexico",
    publisher = "Association for Computational Linguistics",
    url = "https://aclanthology.org/2025.naacl-long.280/",
    doi = "10.18653/v1/2025.naacl-long.280",
    pages = "5420--5443",
    ISBN = "979-8-89176-189-6"
}

\newpage

\appendix

\section{Models used in the study}
\label{sec:models}

A summary of the LLMs included in our study is given in \Fref{tab:models}.
These include three GPT-5 family models\footnote{\url{https://openai.com/index/introducing-gpt-5/}}, two sizes of Gemma 3 models \citep{gemmateam2025gemma3technicalreport}, one Llama-3.1 model \citep{grattafiori2024llama3herdmodels}, Command A \citep{cohere2025commandaenterprisereadylarge}, Mistral NeMo\footnote{\url{https://mistral.ai/news/mistral-nemo}} and Ministral \footnote{\url{https://mistral.ai/news/ministraux/}}, GLM-4 \citep{glm2024chatglm}, Qwen 3 \citep{yang2025qwen3technicalreport}, and Yi-1.5 \citep{ai2025yiopenfoundationmodels}.

\begin{table*}[tb]
    \centering
    \begin{tabular}{lll}
        Model & Size & Origin \\
        \hline
        \texttt{gpt-5.2} & ? & USA \\
        \texttt{gpt-5-mini} & ? & USA \\
        \texttt{gpt-5-nano} & ? & USA \\
        \texttt{gemma-3-4b-it} & 4B & USA \\
        \texttt{gemma-3-27b-it} & 27B & USA \\
        \texttt{Llama-3.1-8B-Instruct} & 8B & USA \\
        \texttt{c4ai-command-r7b-12-2024} & 7B & Canada \\
        \texttt{Ministral-8B-Instruct-2410} & 8B & France \\
        \texttt{Ministral-Nemo-Instruct-2407} & 12B & France \\
        \texttt{GLM-4-9B-0414} & 9B & PRC \\
        \texttt{Qwen3-4B} & 4B & PRC \\
        \texttt{Yi-1.5-9B-32K} & 9B & PRC \\
    \end{tabular}
    \caption{Large Language Models (LLMs) used in this study.}
    \label{tab:models}
\end{table*}

\section{Monolingual haystack retrieval summary}
\label{sec:monolingual-haystacks}

In \Fref{tab:outcomes-monolingual} we show the retrieval rates in monolingual haystacks. This complements \Fref{tab:outcomes} in the main text and demonstrates that results are qualitatively similar to the multilingual setting.

\begin{table*}[t]
\centering
\begin{tabular}{l|rrr|rrr}
 & \multicolumn{3}{c}{1~000 words} & \multicolumn{3}{c}{25~000 words} \\
Model & Both & None & One & Both & None & One \\
\hline
\multicolumn{7}{c}{Monolingual haystacks} \\
\hline
\textsc{gemma-3-27b-it} & 21 & 0 & 579 & 0 & 49 & 551 \\
\textsc{gemma-3-4b-it} & 8 & 7 & 585 & 0 & 189 & 411 \\
\textsc{Llama-3.1-8B-Instruct} & 7 & 15 & 578 & 0 & 6 & 594 \\
\textsc{gpt-5.2-2025-12-11} & 2 & 0 & 598 & 10 & 8 & 582 \\
\textsc{gpt-5-mini-2025-08-07} & 4 & 0 & 596 & 12 & 2 & 586 \\
\textsc{gpt-5-nano-2025-08-07} & 0 & 0 & 600 & 0 & 4 & 596 \\
\textsc{c4ai-command-r7b-12-2024} & 1 & 6 & 593 & 1 & 10 & 589 \\
\textsc{Ministral-8B-Instruct-2410} & 0 & 3 & 597 & 0 & 40 & 560 \\
\textsc{Mistral-Nemo-Instruct-2407} & 0 & 16 & 584 & 0 & 289 & 311 \\
\textsc{GLM-4-9B-0414} & 0 & 3 & 597 & 0 & 171 & 429 \\
\textsc{Qwen3-4B} & 2 & 19 & 579 & 0 & 52 & 548 \\
\textsc{Yi-1.5-9B-32K} & 0 & 29 & 571 & 0 & 327 & 273 \\

\hline
\end{tabular}
\caption{Summary of outcomes from all conflicting haystack retrievals in the monolingual setting. We report the number of times that the model correctly identifies \textbf{both} answers, or retrieval fails so that \textbf{none} of the answers is identified, or reports only \textbf{one} of the two possible answers without mentioning the other.}
\label{tab:outcomes-monolingual}
\end{table*}

\section{Full tables of language bias results}
\label{sec:summary}

Here we present the full tables of results for the largest (25~000 words) haystack size configuration:
\begin{itemize}
    \item Number of wins for each pair of $(L_1, L_2)$ summed over all prompting languages (\Fref{tab:pairs25k}).
    \item Ratio of wins for each $L_1$ summed over all $L_2$, shows separately for each prompting language (\Fref{tab:prompt25k}).
\end{itemize}

\begin{table*}
\begin{minipage}{0.5\textwidth}
\begin{tabular}{l|rrrrr|r}
 & cmn & deu & eng & rus & tur & $\Sigma$ \\
\hline
\multicolumn{7}{c}{\sc gemma-3-27b-it} \\
\hline
cmn &-- & \textcolor{lightgray}{20} & \textbf{35} & \textbf{64} & \textcolor{lightgray}{23} & 142 \\
deu &\textcolor{lightgray}{6} & -- & \textcolor{lightgray}{9} & \textcolor{lightgray}{31} & \textcolor{lightgray}{6} & 52 \\
eng &9 & \textcolor{lightgray}{14} & -- & \textbf{53} & \textcolor{lightgray}{7} & 83 \\
rus &5 & \textcolor{lightgray}{13} & 7 & -- & 12 & 37 \\
tur &\textcolor{lightgray}{11} & \textcolor{lightgray}{15} & \textcolor{lightgray}{14} & \textbf{43} & -- & 83 \\
\hline
\multicolumn{7}{c}{\sc gemma-3-4b-it} \\
\hline
cmn &-- & \textcolor{lightgray}{5} & \textcolor{lightgray}{17} & \textbf{36} & \textcolor{lightgray}{13} & 71 \\
deu &\textcolor{lightgray}{9} & -- & \textcolor{lightgray}{13} & \textbf{26} & \textcolor{lightgray}{12} & 60 \\
eng &\textcolor{lightgray}{6} & \textcolor{lightgray}{7} & -- & \textbf{32} & \textcolor{lightgray}{7} & 52 \\
rus &2 & 2 & 3 & -- & 2 & 9 \\
tur &\textcolor{lightgray}{2} & \textcolor{lightgray}{2} & \textcolor{lightgray}{9} & \textbf{27} & -- & 40 \\
\hline
\multicolumn{7}{c}{\sc Llama-3.1-8B-Instruct} \\
\hline
cmn &-- & \textcolor{lightgray}{24} & \textbf{48} & \textbf{65} & \textcolor{lightgray}{19} & 156 \\
deu &\textcolor{lightgray}{14} & -- & \textbf{18} & \textbf{49} & \textcolor{lightgray}{13} & 94 \\
eng &7 & 3 & -- & \textbf{49} & 5 & 64 \\
rus &16 & 9 & 20 & -- & 6 & 51 \\
tur &\textcolor{lightgray}{11} & \textcolor{lightgray}{23} & \textbf{25} & \textbf{53} & -- & 112 \\
\hline
\multicolumn{7}{c}{\sc gpt-5.2-2025-12-11} \\
\hline
cmn &-- & 11 & 5 & \textcolor{lightgray}{39} & \textcolor{lightgray}{26} & 81 \\
deu &\textbf{38} & -- & 13 & \textbf{60} & \textbf{37} & 148 \\
eng &\textbf{72} & \textbf{50} & -- & \textbf{80} & \textbf{71} & 273 \\
rus &\textcolor{lightgray}{23} & 13 & 6 & -- & \textcolor{lightgray}{25} & 67 \\
tur &\textcolor{lightgray}{27} & 14 & 5 & \textcolor{lightgray}{34} & -- & 80 \\
\hline
\multicolumn{7}{c}{\sc gpt-5-mini-2025-08-07} \\
\hline
cmn &-- & \textcolor{lightgray}{17} & 5 & \textbf{62} & \textcolor{lightgray}{21} & 105 \\
deu &\textcolor{lightgray}{40} & -- & 12 & \textbf{70} & \textcolor{lightgray}{36} & 158 \\
eng &\textbf{51} & \textbf{37} & -- & \textbf{93} & \textbf{40} & 221 \\
rus &14 & 5 & 4 & -- & 15 & 38 \\
tur &\textcolor{lightgray}{26} & \textcolor{lightgray}{15} & 15 & \textbf{57} & -- & 113 \\
\hline
\multicolumn{7}{c}{\sc gpt-5-nano-2025-08-07} \\
\hline
cmn &-- & \textbf{42} & \textbf{54} & \textbf{64} & \textcolor{lightgray}{7} & 167 \\
deu &7 & -- & \textcolor{lightgray}{25} & \textcolor{lightgray}{30} & 5 & 67 \\
eng &11 & \textcolor{lightgray}{19} & -- & \textcolor{lightgray}{19} & 5 & 54 \\
rus &1 & \textcolor{lightgray}{21} & \textcolor{lightgray}{9} & -- & 5 & 36 \\
tur &\textcolor{lightgray}{21} & \textbf{36} & \textbf{45} & \textbf{52} & -- & 154 \\
\hline
\end{tabular}
\end{minipage}
\hfill
\begin{minipage}{0.5\textwidth}
\begin{tabular}{l|rrrrr|r}
 & cmn & deu & eng & rus & tur & $\Sigma$ \\
\hline
\multicolumn{7}{c}{\sc c4ai-command-r7b-12-2024} \\
\hline
cmn &-- & \textcolor{lightgray}{12} & \textcolor{lightgray}{15} & \textcolor{lightgray}{44} & \textcolor{lightgray}{9} & 80 \\
deu &\textcolor{lightgray}{14} & -- & \textcolor{lightgray}{8} & \textcolor{lightgray}{37} & \textcolor{lightgray}{10} & 69 \\
eng &\textcolor{lightgray}{28} & \textcolor{lightgray}{19} & -- & \textcolor{lightgray}{29} & \textcolor{lightgray}{26} & 102 \\
rus &\textcolor{lightgray}{23} & \textcolor{lightgray}{23} & \textcolor{lightgray}{31} & -- & 18 & 95 \\
tur &\textcolor{lightgray}{13} & \textcolor{lightgray}{12} & \textcolor{lightgray}{10} & \textbf{56} & -- & 91 \\
\hline
\multicolumn{7}{c}{\sc Ministral-8B-Instruct-2410} \\
\hline
cmn &-- & \textcolor{lightgray}{19} & \textbf{30} & \textbf{52} & \textcolor{lightgray}{18} & 119 \\
deu &\textcolor{lightgray}{6} & -- & \textcolor{lightgray}{15} & \textbf{33} & \textcolor{lightgray}{4} & 58 \\
eng &6 & \textcolor{lightgray}{3} & -- & \textbf{51} & \textcolor{lightgray}{8} & 68 \\
rus &10 & 8 & 14 & -- & 9 & 41 \\
tur &\textcolor{lightgray}{11} & \textcolor{lightgray}{9} & \textcolor{lightgray}{19} & \textbf{49} & -- & 88 \\
\hline
\multicolumn{7}{c}{\sc Mistral-Nemo-Instruct-2407} \\
\hline
cmn &-- & \textcolor{lightgray}{9} & \textcolor{lightgray}{12} & \textbf{27} & \textcolor{lightgray}{9} & 57 \\
deu &\textcolor{lightgray}{3} & -- & \textcolor{lightgray}{1} & \textcolor{lightgray}{8} & \textcolor{lightgray}{12} & 24 \\
eng &\textcolor{lightgray}{12} & \textcolor{lightgray}{12} & -- & \textcolor{lightgray}{18} & \textcolor{lightgray}{21} & 63 \\
rus &1 & \textcolor{lightgray}{3} & \textcolor{lightgray}{7} & -- & \textcolor{lightgray}{2} & 13 \\
tur &\textcolor{lightgray}{5} & \textcolor{lightgray}{6} & \textcolor{lightgray}{6} & \textcolor{lightgray}{9} & -- & 26 \\
\hline
\multicolumn{7}{c}{\sc GLM-4-9B-0414} \\
\hline
cmn &-- & \textcolor{lightgray}{19} & \textbf{32} & \textbf{49} & \textcolor{lightgray}{15} & 115 \\
deu &\textcolor{lightgray}{4} & -- & \textcolor{lightgray}{18} & \textbf{23} & \textcolor{lightgray}{6} & 51 \\
eng &9 & \textcolor{lightgray}{10} & -- & \textcolor{lightgray}{14} & \textcolor{lightgray}{6} & 39 \\
rus &2 & 4 & \textcolor{lightgray}{6} & -- & 0 & 12 \\
tur &\textcolor{lightgray}{6} & \textcolor{lightgray}{9} & \textcolor{lightgray}{22} & \textbf{18} & -- & 55 \\
\hline
\multicolumn{7}{c}{\sc Qwen3-4B} \\
\hline
cmn &-- & \textcolor{lightgray}{11} & \textcolor{lightgray}{42} & \textbf{49} & \textcolor{lightgray}{7} & 109 \\
deu &\textcolor{lightgray}{19} & -- & \textcolor{lightgray}{19} & \textbf{48} & \textcolor{lightgray}{20} & 106 \\
eng &\textcolor{lightgray}{18} & \textcolor{lightgray}{6} & -- & \textbf{33} & \textcolor{lightgray}{16} & 73 \\
rus &1 & 0 & 3 & -- & 0 & 4 \\
tur &\textcolor{lightgray}{12} & \textcolor{lightgray}{15} & \textcolor{lightgray}{19} & \textbf{51} & -- & 97 \\
\hline
\multicolumn{7}{c}{\sc Yi-1.5-9B-32K} \\
\hline
cmn &-- & \textcolor{lightgray}{9} & \textbf{22} & \textbf{16} & \textcolor{lightgray}{17} & 64 \\
deu &\textcolor{lightgray}{2} & -- & \textcolor{lightgray}{13} & \textcolor{lightgray}{1} & \textcolor{lightgray}{0} & 16 \\
eng &5 & \textcolor{lightgray}{12} & -- & \textcolor{lightgray}{7} & \textcolor{lightgray}{1} & 25 \\
rus &0 & \textcolor{lightgray}{0} & \textcolor{lightgray}{0} & -- & \textcolor{lightgray}{0} & 0 \\
tur &\textcolor{lightgray}{4} & \textcolor{lightgray}{0} & \textcolor{lightgray}{10} & \textcolor{lightgray}{0} & -- & 14 \\
\hline
\end{tabular}
\end{minipage}
\hfill
\caption{Total number of times, over all prompt languages, where language $L_1$ (row) is chosen over $L_2$ (column) in \emph{both} orders of a contrastive pair. Bold: row language wins over column language. Light gray: non-significant difference (Bonferroni-corrected $p < 0.05$). Haystack size is 25~000.}
\label{tab:pairs25k}
\end{table*}

\begin{table*}
\begin{minipage}{0.5\textwidth}
\begin{tabular}{l|rrrrr|r}
 & cmn & deu & eng & rus & tur & Mean \\
\hline
\multicolumn{7}{c}{\sc gemma-3-27b-it} \\
\hline
cmn &\textbf{92} & 85 & 76 & \textcolor{lightgray}{60} & 88 & 80 \\
deu &\textcolor{lightgray}{30} & \textcolor{lightgray}{\textbf{64}} & \textcolor{lightgray}{60} & 10 & \textcolor{lightgray}{52} & 43 \\
eng &\textcolor{lightgray}{52} & \textcolor{lightgray}{63} & \textcolor{lightgray}{\textbf{69}} & \textcolor{lightgray}{39} & \textcolor{lightgray}{50} & 54 \\
rus &8 & 4 & 0 & \textbf{84} & 0 & 19 \\
tur &\textcolor{lightgray}{60} & \textcolor{lightgray}{57} & \textcolor{lightgray}{72} & \textcolor{lightgray}{30} & \textbf{87} & 61 \\
\hline
\multicolumn{7}{c}{\sc gemma-3-4b-it} \\
\hline
cmn &\textbf{92} & \textcolor{lightgray}{67} & \textcolor{lightgray}{73} & \textcolor{lightgray}{60} & 94 & 77 \\
deu &\textcolor{lightgray}{71} & \textbf{95} & 90 & \textcolor{lightgray}{54} & \textcolor{lightgray}{70} & 76 \\
eng &\textcolor{lightgray}{40} & \textcolor{lightgray}{59} & \textcolor{lightgray}{\textbf{77}} & \textcolor{lightgray}{38} & \textcolor{lightgray}{53} & 54 \\
rus &0 & 0 & 0 & \textcolor{lightgray}{\textbf{56}} & 0 & 11 \\
tur &\textcolor{lightgray}{58} & \textcolor{lightgray}{53} & \textcolor{lightgray}{54} & \textcolor{lightgray}{38} & \textcolor{lightgray}{\textbf{64}} & 53 \\
\hline
\multicolumn{7}{c}{\sc Llama-3.1-8B-Instruct} \\
\hline
cmn &\textbf{98} & 77 & 78 & \textcolor{lightgray}{52} & 78 & 77 \\
deu &\textcolor{lightgray}{56} & \textbf{86} & \textcolor{lightgray}{75} & \textcolor{lightgray}{26} & \textcolor{lightgray}{63} & 61 \\
eng &\textcolor{lightgray}{38} & \textcolor{lightgray}{48} & \textcolor{lightgray}{\textbf{58}} & 11 & \textcolor{lightgray}{42} & 39 \\
rus &2 & 2 & 2 & \textbf{96} & 0 & 20 \\
tur &\textcolor{lightgray}{74} & \textcolor{lightgray}{69} & \textcolor{lightgray}{67} & \textcolor{lightgray}{52} & \textbf{94} & 71 \\
\hline
\multicolumn{7}{c}{\sc gpt-5.2-2025-12-11} \\
\hline
cmn &\textcolor{lightgray}{\textbf{69}} & 26 & \textcolor{lightgray}{33} & 20 & 22 & 34 \\
deu &\textcolor{lightgray}{58} & \textbf{82} & \textcolor{lightgray}{68} & \textcolor{lightgray}{48} & \textcolor{lightgray}{56} & 62 \\
eng &88 & 90 & \textbf{97} & 89 & 87 & 90 \\
rus &9 & 15 & 20 & \textcolor{lightgray}{\textbf{67}} & 11 & 24 \\
tur &\textcolor{lightgray}{28} & 25 & 23 & 20 & \textcolor{lightgray}{\textbf{70}} & 33 \\
\hline
\multicolumn{7}{c}{\sc gpt-5-mini-2025-08-07} \\
\hline
cmn &\textbf{80} & \textcolor{lightgray}{40} & \textcolor{lightgray}{42} & 19 & \textcolor{lightgray}{40} & 44 \\
deu &\textcolor{lightgray}{61} & \textbf{94} & \textcolor{lightgray}{67} & \textcolor{lightgray}{57} & \textcolor{lightgray}{57} & 67 \\
eng &92 & 87 & \textbf{95} & 84 & 73 & 86 \\
rus &0 & 5 & 3 & \textcolor{lightgray}{\textbf{60}} & 3 & 14 \\
tur &\textcolor{lightgray}{40} & \textcolor{lightgray}{38} & \textcolor{lightgray}{55} & 23 & \textbf{92} & 49 \\
\hline
\multicolumn{7}{c}{\sc gpt-5-nano-2025-08-07} \\
\hline
cmn &\textbf{92} & \textcolor{lightgray}{73} & \textcolor{lightgray}{74} & 77 & 84 & 80 \\
deu &\textcolor{lightgray}{33} & \textcolor{lightgray}{\textbf{57}} & \textcolor{lightgray}{41} & \textcolor{lightgray}{31} & 18 & 36 \\
eng &21 & \textcolor{lightgray}{28} & \textcolor{lightgray}{\textbf{40}} & 23 & \textcolor{lightgray}{31} & 29 \\
rus &5 & 19 & 17 & \textcolor{lightgray}{\textbf{36}} & 14 & 18 \\
tur &90 & 76 & 85 & 87 & \textbf{100} & 88 \\
\hline
\end{tabular}
\end{minipage}
\hfill
\begin{minipage}{0.5\textwidth}
\begin{tabular}{l|rrrrr|r}
 & cmn & deu & eng & rus & tur & Mean \\
\hline
\multicolumn{7}{c}{\sc c4ai-command-r7b-12-2024} \\
\hline
cmn &\textbf{97} & \textcolor{lightgray}{31} & \textcolor{lightgray}{32} & 13 & \textcolor{lightgray}{70} & 49 \\
deu &\textcolor{lightgray}{42} & \textcolor{lightgray}{\textbf{77}} & \textcolor{lightgray}{65} & 14 & \textcolor{lightgray}{44} & 48 \\
eng &\textcolor{lightgray}{31} & \textcolor{lightgray}{75} & \textbf{100} & \textcolor{lightgray}{48} & \textcolor{lightgray}{40} & 59 \\
rus &\textcolor{lightgray}{27} & 22 & 2 & \textbf{100} & 24 & 35 \\
tur &\textcolor{lightgray}{62} & \textcolor{lightgray}{52} & \textcolor{lightgray}{66} & 19 & \textbf{86} & 57 \\
\hline
\multicolumn{7}{c}{\sc Ministral-8B-Instruct-2410} \\
\hline
cmn &\textbf{100} & \textcolor{lightgray}{68} & \textcolor{lightgray}{79} & \textcolor{lightgray}{50} & \textcolor{lightgray}{71} & 74 \\
deu &\textcolor{lightgray}{48} & \textbf{92} & \textcolor{lightgray}{63} & \textcolor{lightgray}{33} & \textcolor{lightgray}{50} & 57 \\
eng &\textcolor{lightgray}{34} & \textcolor{lightgray}{64} & \textbf{82} & 8 & \textcolor{lightgray}{40} & 46 \\
rus &2 & 2 & 0 & \textbf{89} & 0 & 19 \\
tur &\textcolor{lightgray}{64} & \textcolor{lightgray}{62} & \textcolor{lightgray}{64} & \textcolor{lightgray}{32} & \textbf{100} & 64 \\
\hline
\multicolumn{7}{c}{\sc Mistral-Nemo-Instruct-2407} \\
\hline
cmn &\textbf{100} & \textcolor{lightgray}{70} & \textcolor{lightgray}{53} & \textcolor{lightgray}{64} & \textcolor{lightgray}{67} & 71 \\
deu &\textcolor{lightgray}{38} & \textcolor{lightgray}{\textbf{67}} & \textcolor{lightgray}{47} & \textcolor{lightgray}{27} & \textcolor{lightgray}{20} & 40 \\
eng &\textcolor{lightgray}{50} & \textcolor{lightgray}{85} & \textbf{96} & \textcolor{lightgray}{40} & \textcolor{lightgray}{62} & 67 \\
rus &0 & 0 & 0 & \textcolor{lightgray}{\textbf{87}} & \textcolor{lightgray}{0} & 17 \\
tur &\textcolor{lightgray}{25} & \textcolor{lightgray}{31} & \textcolor{lightgray}{30} & \textcolor{lightgray}{29} & \textcolor{lightgray}{\textbf{82}} & 39 \\
\hline
\multicolumn{7}{c}{\sc GLM-4-9B-0414} \\
\hline
cmn &\textbf{94} & 86 & \textcolor{lightgray}{73} & \textcolor{lightgray}{71} & 96 & 84 \\
deu &\textcolor{lightgray}{35} & \textcolor{lightgray}{\textbf{71}} & \textcolor{lightgray}{64} & \textcolor{lightgray}{65} & \textcolor{lightgray}{12} & 49 \\
eng &\textcolor{lightgray}{24} & \textcolor{lightgray}{33} & \textcolor{lightgray}{\textbf{58}} & \textcolor{lightgray}{23} & \textcolor{lightgray}{15} & 31 \\
rus &12 & 5 & 0 & \textcolor{lightgray}{\textbf{31}} & 0 & 10 \\
tur &\textcolor{lightgray}{71} & \textcolor{lightgray}{100} & \textcolor{lightgray}{61} & \textcolor{lightgray}{65} & \textcolor{lightgray}{\textbf{75}} & 74 \\
\hline
\multicolumn{7}{c}{\sc Qwen3-4B} \\
\hline
cmn &\textbf{92} & \textcolor{lightgray}{58} & \textcolor{lightgray}{51} & \textcolor{lightgray}{48} & 86 & 67 \\
deu &\textcolor{lightgray}{61} & \textbf{93} & \textcolor{lightgray}{78} & \textcolor{lightgray}{77} & \textcolor{lightgray}{73} & 76 \\
eng &\textcolor{lightgray}{27} & \textcolor{lightgray}{57} & \textcolor{lightgray}{\textbf{71}} & \textcolor{lightgray}{40} & \textcolor{lightgray}{33} & 46 \\
rus &0 & 0 & 2 & \textcolor{lightgray}{\textbf{16}} & 0 & 4 \\
tur &\textcolor{lightgray}{72} & \textcolor{lightgray}{57} & \textcolor{lightgray}{67} & \textcolor{lightgray}{60} & \textbf{87} & 68 \\
\hline
\multicolumn{7}{c}{\sc Yi-1.5-9B-32K} \\
\hline
cmn &\textbf{95} & \textcolor{lightgray}{71} & \textcolor{lightgray}{75} & 100 & \textcolor{lightgray}{83} & 85 \\
deu &\textcolor{lightgray}{33} & \textcolor{lightgray}{\textbf{88}} & \textcolor{lightgray}{33} & \textcolor{lightgray}{29} & \textcolor{lightgray}{29} & 42 \\
eng &\textcolor{lightgray}{29} & \textcolor{lightgray}{29} & \textcolor{lightgray}{\textbf{78}} & \textcolor{lightgray}{27} & \textcolor{lightgray}{32} & 39 \\
rus &\textcolor{lightgray}{0} & \textcolor{lightgray}{0} & \textcolor{lightgray}{0} & \textcolor{lightgray}{\textbf{0}} & \textcolor{lightgray}{0} & 0 \\
tur &\textcolor{lightgray}{38} & \textcolor{lightgray}{50} & \textcolor{lightgray}{20} & \textcolor{lightgray}{0} & \textcolor{lightgray}{\textbf{70}} & 36 \\
\hline
\end{tabular}
\end{minipage}
\hfill
\caption{Percentage of times, over all contrastive pairs and  languages $L_2$, where language $L_1$ (row) is chosen over $L_2$ in \emph{both} orders of a contrastive pair, when prompting using $L_p$ (column). Bold marks the diagonal, and thus represents the proportion of times when the information in the prompting language is chosen. Black marks significant difference from 50\%, light gray non-significant difference (Bonferroni-corrected $p < 0.05$). Haystack size is 25~000.}
\label{tab:prompt25k}
\end{table*}

\section{Results with reduced prompt}
\label{app:outcomes-short-prompt}

\begin{table*}[!t]
\centering
\begin{tabular}{l|r|rrr}
Model & Haystack & Both & None & One \\
\hline
\textsc{gpt-5.2-2025-12-11} &  1000 & 1113 & 0 & 1887 \\
\textsc{gpt-5-nano-2025-08-07} &  1000 & 367 & 1 & 2632 \\
\textsc{gpt-5-nano-2025-08-07} &  25000 & 12 & 16 & 2972 \\
\hline
\end{tabular}
\caption{Summary of outcomes from all conflicting haystack retrievals.}
\label{tab:short-prompt}
\end{table*}

\Fref{tab:short-prompt} presents outcome statistics for a selection of representative configurations using a prompt that omits the final sentence (``Answer with only the full name of [person].'' and its translations into the other languages). The intention is to show that even state-of-the-art models in the easiest setting (short haystack) confidently assert a single answer in the face of contradictory information.

\section{Additional language bias values}
\label{app:biasparam}

Here we present the estimates of $b_{{l_1},{l_2}}$ for intermediate haystack sizes: 2500 words (\Fref{tab:biasparam2k5}, 5000 words (\Fref{tab:biasparam5k}, and 10000 words (\Fref{tab:biasparam10k}. The estimates for the smallest (1000) and largest (25~000) are found in the main text.

\begin{table}[t]
\begin{tabular}{lrl}
Language pair $l_1, l_2$ & $P(> 0)$ & 95\% CI \\
\hline
German vs Chinese & 59.8\% & $[-0.62, 0.79]$ \\
English vs Chinese & 68.6\% & $[-0.54, 0.86]$ \\
Chinese vs Russian & 100.0\% & $[2.26, 3.71]$ \\
Chinese vs Turkish & 61.1\% & $[-0.61, 0.76]$ \\
English vs German & 88.1\% & $[-0.30, 1.15]$ \\
German vs Russian & 100.0\% & $[2.42, 3.92]$ \\
German vs Turkish & 93.5\% & $[-0.15, 1.20]$ \\
English vs Russian & 100.0\% & $[2.43, 3.88]$ \\
English vs Turkish & 99.3\% & $[0.14, 1.48]$ \\
Turkish vs Russian & 100.0\% & $[2.52, 4.05]$ \\
\end{tabular}
\caption{Estimates of the log-odds bias parameter $b_{l_1, l_2}$, representing
the overall bias (over all prompting languages and models) of language $l_1$
over language $l_2$. Haystack size is 2500.}
\label{tab:biasparam2k5}
\end{table}

\begin{table}
\begin{tabular}{lrl}
Language pair $l_1, l_2$ & $P(> 0)$ & 95\% CI \\
\hline
German vs Chinese & 75.6\% & $[-0.54, 1.06]$ \\
English vs Chinese & 50.4\% & $[-0.76, 0.78]$ \\
Chinese vs Russian & 100.0\% & $[2.33, 3.96]$ \\
Chinese vs Turkish & 64.8\% & $[-0.63, 0.94]$ \\
English vs German & 67.5\% & $[-0.61, 0.98]$ \\
German vs Russian & 100.0\% & $[2.99, 4.82]$ \\
German vs Turkish & 95.6\% & $[-0.09, 1.52]$ \\
English vs Russian & 100.0\% & $[2.43, 4.18]$ \\
English vs Turkish & 97.1\% & $[-0.03, 1.49]$ \\
Turkish vs Russian & 100.0\% & $[2.17, 3.80]$ \\
\end{tabular}
\caption{Estimates of the log-odds bias parameter $b_{l_1, l_2}$, representing
the overall bias (over all prompting languages and models) of language $l_1$
over language $l_2$. Haystack size is 5000.}
\label{tab:biasparam5k}
\end{table}

\begin{table}
\begin{tabular}{lrl}
Language pair $l_1, l_2$ & $P(> 0)$ & 95\% CI \\
\hline
German vs Chinese & 72.5\% & $[-0.51, 0.97]$ \\
Chinese vs English & 74.6\% & $[-0.49, 0.98]$ \\
Chinese vs Russian & 100.0\% & $[2.61, 4.32]$ \\
Chinese vs Turkish & 69.9\% & $[-0.56, 0.99]$ \\
German vs English & 75.4\% & $[-0.51, 1.06]$ \\
German vs Russian & 100.0\% & $[2.48, 4.21]$ \\
German vs Turkish & 78.0\% & $[-0.46, 1.09]$ \\
English vs Russian & 100.0\% & $[2.60, 4.32]$ \\
English vs Turkish & 51.8\% & $[-0.73, 0.74]$ \\
Turkish vs Russian & 100.0\% & $[2.66, 4.33]$ \\
\end{tabular}
\caption{Estimates of the log-odds bias parameter $b_{l_1, l_2}$, representing
the overall bias (over all prompting languages and models) of language $l_1$
over language $l_2$. Haystack size is 10000.}
\label{tab:biasparam10k}
\end{table}

\section{Additional pairwise bias tables}
\label{app:pairwise}

Here we present tables of the number of wins for each model and language pair, for intermediate haystack sizes: 1000 words (\Fref{tab:pairs1k}), 2500 words (\Fref{tab:pairs2k5}), 5000 words (\Fref{tab:pairs5k}), 10000 words (\Fref{tab:pairs10k}).

\begin{table*}[t]
\begin{minipage}{0.5\textwidth}
\begin{tabular}{l|rrrrr|r}
 & cmn & deu & eng & rus & tur & $\Sigma$ \\
\hline
\multicolumn{7}{c}{\sc gemma-3-27b-it} \\
\hline
cmn &-- & \textcolor{lightgray}{5} & \textcolor{lightgray}{2} & \textbf{39} & \textcolor{lightgray}{3} & 49 \\
deu &\textcolor{lightgray}{11} & -- & \textcolor{lightgray}{2} & \textcolor{lightgray}{16} & \textcolor{lightgray}{9} & 38 \\
eng &\textcolor{lightgray}{4} & \textcolor{lightgray}{6} & -- & \textbf{34} & \textcolor{lightgray}{10} & 54 \\
rus &2 & \textcolor{lightgray}{3} & 2 & -- & \textcolor{lightgray}{5} & 12 \\
tur &\textcolor{lightgray}{17} & \textcolor{lightgray}{4} & \textcolor{lightgray}{3} & \textcolor{lightgray}{19} & -- & 43 \\
\hline
\multicolumn{7}{c}{\sc gemma-3-4b-it} \\
\hline
cmn &-- & \textcolor{lightgray}{20} & \textbf{46} & \textbf{76} & \textcolor{lightgray}{30} & 172 \\
deu &\textcolor{lightgray}{11} & -- & \textcolor{lightgray}{13} & \textbf{78} & \textcolor{lightgray}{30} & 132 \\
eng &6 & \textcolor{lightgray}{18} & -- & \textbf{71} & \textcolor{lightgray}{25} & 120 \\
rus &9 & 2 & 4 & -- & 7 & 22 \\
tur &\textcolor{lightgray}{10} & \textcolor{lightgray}{17} & \textcolor{lightgray}{31} & \textbf{75} & -- & 133 \\
\hline
\multicolumn{7}{c}{\sc Llama-3.1-8B-Instruct} \\
\hline
cmn &-- & \textcolor{lightgray}{5} & \textcolor{lightgray}{8} & \textbf{43} & \textcolor{lightgray}{9} & 65 \\
deu &\textcolor{lightgray}{6} & -- & \textcolor{lightgray}{4} & \textbf{41} & \textcolor{lightgray}{5} & 56 \\
eng &\textcolor{lightgray}{10} & \textcolor{lightgray}{4} & -- & \textbf{51} & \textcolor{lightgray}{9} & 74 \\
rus &12 & 9 & 13 & -- & 18 & 52 \\
tur &\textcolor{lightgray}{6} & \textcolor{lightgray}{0} & \textcolor{lightgray}{1} & \textbf{61} & -- & 68 \\
\hline
\multicolumn{7}{c}{\sc gpt-5.2-2025-12-11} \\
\hline
cmn &-- & 7 & 8 & \textcolor{lightgray}{30} & \textcolor{lightgray}{22} & 67 \\
deu &\textbf{59} & -- & 11 & \textbf{57} & \textbf{48} & 175 \\
eng &\textbf{70} & \textbf{49} & -- & \textbf{76} & \textbf{60} & 255 \\
rus &\textcolor{lightgray}{26} & 10 & 5 & -- & \textcolor{lightgray}{20} & 61 \\
tur &\textcolor{lightgray}{34} & 16 & 5 & \textcolor{lightgray}{40} & -- & 95 \\
\hline
\multicolumn{7}{c}{\sc gpt-5-mini-2025-08-07} \\
\hline
cmn &-- & \textcolor{lightgray}{24} & 16 & \textbf{81} & \textcolor{lightgray}{26} & 147 \\
deu &\textcolor{lightgray}{40} & -- & \textcolor{lightgray}{18} & \textbf{75} & \textcolor{lightgray}{39} & 172 \\
eng &\textbf{54} & \textcolor{lightgray}{39} & -- & \textbf{101} & \textbf{52} & 246 \\
rus &7 & 7 & 3 & -- & 15 & 32 \\
tur &\textcolor{lightgray}{34} & \textcolor{lightgray}{20} & 9 & \textbf{59} & -- & 122 \\
\hline
\multicolumn{7}{c}{\sc gpt-5-nano-2025-08-07} \\
\hline
cmn &-- & \textbf{45} & \textcolor{lightgray}{44} & \textbf{60} & \textcolor{lightgray}{23} & 172 \\
deu &15 & -- & \textcolor{lightgray}{16} & \textbf{48} & 6 & 85 \\
eng &\textcolor{lightgray}{22} & \textcolor{lightgray}{30} & -- & \textbf{59} & \textcolor{lightgray}{21} & 132 \\
rus &11 & 13 & 17 & -- & 7 & 48 \\
tur &\textcolor{lightgray}{21} & \textbf{50} & \textcolor{lightgray}{46} & \textbf{64} & -- & 181 \\
\hline
\end{tabular}
\end{minipage}
\hfill
\begin{minipage}{0.5\textwidth}
\begin{tabular}{l|rrrrr|r}
 & cmn & deu & eng & rus & tur & $\Sigma$ \\
\hline
\multicolumn{7}{c}{\sc c4ai-command-r7b-12-2024} \\
\hline
cmn &-- & \textcolor{lightgray}{15} & 11 & \textcolor{lightgray}{46} & \textbf{35} & 107 \\
deu &\textcolor{lightgray}{14} & -- & 1 & \textcolor{lightgray}{41} & \textcolor{lightgray}{18} & 74 \\
eng &\textbf{40} & \textbf{31} & -- & \textcolor{lightgray}{35} & \textbf{47} & 153 \\
rus &\textcolor{lightgray}{31} & \textcolor{lightgray}{21} & \textcolor{lightgray}{33} & -- & \textcolor{lightgray}{20} & 105 \\
tur &4 & \textcolor{lightgray}{10} & 1 & \textcolor{lightgray}{44} & -- & 59 \\
\hline
\multicolumn{7}{c}{\sc Ministral-8B-Instruct-2410} \\
\hline
cmn &-- & \textbf{14} & \textcolor{lightgray}{13} & \textbf{59} & \textcolor{lightgray}{11} & 97 \\
deu &1 & -- & \textcolor{lightgray}{1} & \textbf{54} & \textcolor{lightgray}{17} & 73 \\
eng &\textcolor{lightgray}{10} & \textcolor{lightgray}{11} & -- & \textbf{58} & \textcolor{lightgray}{24} & 103 \\
rus &3 & 2 & 3 & -- & 13 & 21 \\
tur &\textcolor{lightgray}{3} & \textcolor{lightgray}{13} & \textcolor{lightgray}{7} & \textbf{42} & -- & 65 \\
\hline
\multicolumn{7}{c}{\sc Mistral-Nemo-Instruct-2407} \\
\hline
cmn &-- & \textbf{32} & \textcolor{lightgray}{30} & \textbf{86} & \textcolor{lightgray}{21} & 169 \\
deu &6 & -- & 1 & \textbf{56} & \textcolor{lightgray}{21} & 84 \\
eng &\textcolor{lightgray}{17} & \textbf{18} & -- & \textbf{75} & \textcolor{lightgray}{34} & 144 \\
rus &1 & 4 & 1 & -- & 2 & 8 \\
tur &\textcolor{lightgray}{12} & \textcolor{lightgray}{16} & \textcolor{lightgray}{13} & \textbf{80} & -- & 121 \\
\hline
\multicolumn{7}{c}{\sc GLM-4-9B-0414} \\
\hline
cmn &-- & \textbf{32} & \textcolor{lightgray}{24} & \textbf{84} & \textcolor{lightgray}{27} & 167 \\
deu &8 & -- & \textcolor{lightgray}{5} & \textbf{61} & \textcolor{lightgray}{14} & 88 \\
eng &\textcolor{lightgray}{7} & \textcolor{lightgray}{9} & -- & \textbf{79} & \textcolor{lightgray}{10} & 105 \\
rus &7 & 10 & 15 & -- & 3 & 35 \\
tur &\textcolor{lightgray}{11} & \textcolor{lightgray}{14} & \textcolor{lightgray}{25} & \textbf{70} & -- & 120 \\
\hline
\multicolumn{7}{c}{\sc Qwen3-4B} \\
\hline
cmn &-- & \textcolor{lightgray}{17} & \textcolor{lightgray}{25} & \textbf{84} & \textcolor{lightgray}{16} & 142 \\
deu &\textcolor{lightgray}{16} & -- & \textcolor{lightgray}{12} & \textbf{62} & \textcolor{lightgray}{20} & 110 \\
eng &\textcolor{lightgray}{10} & \textcolor{lightgray}{21} & -- & \textbf{83} & \textcolor{lightgray}{22} & 136 \\
rus &1 & 5 & 4 & -- & 0 & 10 \\
tur &\textcolor{lightgray}{25} & \textcolor{lightgray}{22} & \textcolor{lightgray}{24} & \textbf{83} & -- & 154 \\
\hline
\multicolumn{7}{c}{\sc Yi-1.5-9B-32K} \\
\hline
cmn &-- & \textcolor{lightgray}{31} & \textcolor{lightgray}{19} & \textbf{82} & \textcolor{lightgray}{23} & 155 \\
deu &\textcolor{lightgray}{14} & -- & \textcolor{lightgray}{7} & \textbf{75} & \textcolor{lightgray}{29} & 125 \\
eng &\textcolor{lightgray}{20} & \textcolor{lightgray}{13} & -- & \textbf{72} & \textbf{33} & 138 \\
rus &3 & 7 & 7 & -- & 5 & 22 \\
tur &\textcolor{lightgray}{7} & \textcolor{lightgray}{15} & 8 & \textbf{72} & -- & 102 \\
\hline
\end{tabular}
\end{minipage}
\hfill
\caption{Total number of times, over all prompt languages, where language $L_1$ (row) is chosen over $L_2$ (column) in \emph{both} orders of a contrastive pair. Bold: row language wins over column language. Light gray: non-significant difference (Bonferroni-corrected $p < 0.05$). Haystack size is 1000.}
\label{tab:pairs1k}
\end{table*}

\begin{table*}[t]
\begin{minipage}{0.5\textwidth}
\begin{tabular}{l|rrrrr|r}
 & cmn & deu & eng & rus & tur & $\Sigma$ \\
\hline
\multicolumn{7}{c}{\sc gemma-3-27b-it} \\
\hline
cmn &-- & 2 & \textcolor{lightgray}{8} & \textbf{44} & \textcolor{lightgray}{11} & 65 \\
deu &\textbf{16} & -- & \textcolor{lightgray}{8} & \textcolor{lightgray}{31} & \textcolor{lightgray}{7} & 62 \\
eng &\textcolor{lightgray}{20} & \textcolor{lightgray}{11} & -- & \textbf{51} & \textcolor{lightgray}{15} & 97 \\
rus &13 & \textcolor{lightgray}{13} & 4 & -- & 12 & 42 \\
tur &\textcolor{lightgray}{15} & \textcolor{lightgray}{7} & \textcolor{lightgray}{12} & \textbf{34} & -- & 68 \\
\hline
\multicolumn{7}{c}{\sc gemma-3-4b-it} \\
\hline
cmn &-- & \textcolor{lightgray}{23} & \textcolor{lightgray}{26} & \textbf{78} & \textbf{33} & 160 \\
deu &\textcolor{lightgray}{18} & -- & \textcolor{lightgray}{28} & \textbf{63} & \textbf{28} & 137 \\
eng &\textcolor{lightgray}{18} & \textcolor{lightgray}{12} & -- & \textbf{72} & \textcolor{lightgray}{30} & 132 \\
rus &11 & 7 & 10 & -- & 11 & 39 \\
tur &4 & 5 & \textcolor{lightgray}{15} & \textbf{63} & -- & 87 \\
\hline
\multicolumn{7}{c}{\sc Llama-3.1-8B-Instruct} \\
\hline
cmn &-- & \textcolor{lightgray}{14} & \textcolor{lightgray}{14} & \textbf{62} & \textcolor{lightgray}{7} & 97 \\
deu &\textcolor{lightgray}{12} & -- & \textcolor{lightgray}{14} & \textbf{69} & \textcolor{lightgray}{9} & 104 \\
eng &\textcolor{lightgray}{5} & \textcolor{lightgray}{7} & -- & \textbf{62} & \textcolor{lightgray}{12} & 86 \\
rus &17 & 13 & 21 & -- & 13 & 64 \\
tur &\textcolor{lightgray}{11} & \textcolor{lightgray}{8} & \textcolor{lightgray}{8} & \textbf{68} & -- & 95 \\
\hline
\multicolumn{7}{c}{\sc gpt-5.2-2025-12-11} \\
\hline
cmn &-- & 9 & 2 & \textcolor{lightgray}{34} & 13 & 58 \\
deu &\textbf{56} & -- & 7 & \textbf{52} & \textbf{47} & 162 \\
eng &\textbf{71} & \textbf{54} & -- & \textbf{79} & \textbf{69} & 273 \\
rus &\textcolor{lightgray}{20} & 12 & 3 & -- & \textcolor{lightgray}{23} & 58 \\
tur &\textbf{41} & 12 & 7 & \textcolor{lightgray}{37} & -- & 97 \\
\hline
\multicolumn{7}{c}{\sc gpt-5-mini-2025-08-07} \\
\hline
cmn &-- & 18 & 13 & \textbf{70} & 20 & 121 \\
deu &\textbf{55} & -- & \textcolor{lightgray}{21} & \textbf{85} & \textbf{44} & 205 \\
eng &\textbf{64} & \textcolor{lightgray}{32} & -- & \textbf{96} & \textbf{52} & 244 \\
rus &11 & 6 & 5 & -- & 14 & 36 \\
tur &\textbf{47} & 14 & 10 & \textbf{64} & -- & 135 \\
\hline
\multicolumn{7}{c}{\sc gpt-5-nano-2025-08-07} \\
\hline
cmn &-- & \textbf{29} & \textbf{42} & \textbf{52} & \textcolor{lightgray}{7} & 130 \\
deu &9 & -- & \textcolor{lightgray}{35} & \textbf{41} & 4 & 89 \\
eng &16 & \textcolor{lightgray}{34} & -- & \textbf{52} & 11 & 113 \\
rus &5 & 12 & 12 & -- & 0 & 29 \\
tur &\textcolor{lightgray}{19} & \textbf{33} & \textbf{58} & \textbf{60} & -- & 170 \\
\hline
\end{tabular}
\end{minipage}
\hfill
\begin{minipage}{0.5\textwidth}
\begin{tabular}{l|rrrrr|r}
 & cmn & deu & eng & rus & tur & $\Sigma$ \\
\hline
\multicolumn{7}{c}{\sc c4ai-command-r7b-12-2024} \\
\hline
cmn &-- & \textcolor{lightgray}{15} & 8 & \textcolor{lightgray}{39} & \textcolor{lightgray}{15} & 77 \\
deu &\textcolor{lightgray}{20} & -- & 0 & \textcolor{lightgray}{45} & \textcolor{lightgray}{20} & 85 \\
eng &\textbf{42} & \textbf{48} & -- & \textcolor{lightgray}{41} & \textbf{53} & 184 \\
rus &\textcolor{lightgray}{39} & \textcolor{lightgray}{34} & \textcolor{lightgray}{36} & -- & 30 & 139 \\
tur &\textcolor{lightgray}{17} & \textcolor{lightgray}{9} & 1 & \textbf{61} & -- & 88 \\
\hline
\multicolumn{7}{c}{\sc Ministral-8B-Instruct-2410} \\
\hline
cmn &-- & \textcolor{lightgray}{15} & \textcolor{lightgray}{11} & \textbf{67} & \textcolor{lightgray}{14} & 107 \\
deu &\textcolor{lightgray}{15} & -- & \textcolor{lightgray}{2} & \textbf{66} & \textcolor{lightgray}{15} & 98 \\
eng &\textcolor{lightgray}{14} & \textcolor{lightgray}{5} & -- & \textbf{69} & \textcolor{lightgray}{11} & 99 \\
rus &4 & 7 & 8 & -- & 9 & 28 \\
tur &\textcolor{lightgray}{8} & \textcolor{lightgray}{15} & \textcolor{lightgray}{17} & \textbf{58} & -- & 98 \\
\hline
\multicolumn{7}{c}{\sc Mistral-Nemo-Instruct-2407} \\
\hline
cmn &-- & \textbf{29} & \textcolor{lightgray}{32} & \textbf{80} & \textcolor{lightgray}{15} & 156 \\
deu &7 & -- & \textcolor{lightgray}{2} & \textbf{64} & \textcolor{lightgray}{15} & 88 \\
eng &\textcolor{lightgray}{19} & \textcolor{lightgray}{11} & -- & \textbf{60} & \textcolor{lightgray}{15} & 105 \\
rus &1 & 5 & 3 & -- & 4 & 13 \\
tur &\textcolor{lightgray}{12} & \textcolor{lightgray}{22} & \textcolor{lightgray}{19} & \textbf{62} & -- & 115 \\
\hline
\multicolumn{7}{c}{\sc GLM-4-9B-0414} \\
\hline
cmn &-- & \textbf{30} & \textbf{26} & \textbf{77} & \textbf{34} & 167 \\
deu &6 & -- & \textcolor{lightgray}{20} & \textbf{66} & \textcolor{lightgray}{26} & 118 \\
eng &0 & \textcolor{lightgray}{11} & -- & \textbf{69} & \textcolor{lightgray}{18} & 98 \\
rus &4 & 1 & 13 & -- & 2 & 20 \\
tur &9 & \textcolor{lightgray}{19} & \textcolor{lightgray}{10} & \textbf{69} & -- & 107 \\
\hline
\multicolumn{7}{c}{\sc Qwen3-4B} \\
\hline
cmn &-- & \textcolor{lightgray}{13} & \textcolor{lightgray}{24} & \textbf{73} & \textcolor{lightgray}{13} & 123 \\
deu &\textcolor{lightgray}{21} & -- & \textcolor{lightgray}{23} & \textbf{63} & \textcolor{lightgray}{27} & 134 \\
eng &\textcolor{lightgray}{13} & \textcolor{lightgray}{9} & -- & \textbf{70} & \textcolor{lightgray}{16} & 108 \\
rus &6 & 2 & 5 & -- & 2 & 15 \\
tur &\textcolor{lightgray}{29} & \textcolor{lightgray}{24} & \textcolor{lightgray}{29} & \textbf{75} & -- & 157 \\
\hline
\multicolumn{7}{c}{\sc Yi-1.5-9B-32K} \\
\hline
cmn &-- & \textcolor{lightgray}{21} & \textcolor{lightgray}{27} & \textbf{79} & \textbf{45} & 172 \\
deu &\textcolor{lightgray}{20} & -- & \textcolor{lightgray}{24} & \textbf{76} & \textcolor{lightgray}{29} & 149 \\
eng &\textcolor{lightgray}{22} & \textcolor{lightgray}{11} & -- & \textbf{72} & \textcolor{lightgray}{36} & 141 \\
rus &2 & 5 & 8 & -- & 3 & 18 \\
tur &6 & \textcolor{lightgray}{22} & \textcolor{lightgray}{22} & \textbf{80} & -- & 130 \\
\hline
\end{tabular}
\end{minipage}
\hfill
\caption{Total number of times, over all prompt languages, where language $L_1$ (row) is chosen over $L_2$ (column) in \emph{both} orders of a contrastive pair. Bold: row language wins over column language. Light gray: non-significant difference (Bonferroni-corrected $p < 0.05$). Haystack size is 2500.}
\label{tab:pairs2k5}
\end{table*}

\begin{table*}[t]
\begin{minipage}{0.5\textwidth}
\begin{tabular}{l|rrrrr|r}
 & cmn & deu & eng & rus & tur & $\Sigma$ \\
\hline
\multicolumn{7}{c}{\sc gemma-3-27b-it} \\
\hline
cmn &-- & \textcolor{lightgray}{9} & \textcolor{lightgray}{8} & \textbf{41} & \textcolor{lightgray}{12} & 70 \\
deu &\textcolor{lightgray}{11} & -- & \textcolor{lightgray}{13} & \textbf{37} & \textcolor{lightgray}{19} & 80 \\
eng &\textcolor{lightgray}{24} & \textcolor{lightgray}{18} & -- & \textbf{37} & \textcolor{lightgray}{18} & 97 \\
rus &8 & 11 & 6 & -- & \textcolor{lightgray}{21} & 46 \\
tur &\textcolor{lightgray}{8} & \textcolor{lightgray}{6} & \textcolor{lightgray}{10} & \textcolor{lightgray}{31} & -- & 55 \\
\hline
\multicolumn{7}{c}{\sc gemma-3-4b-it} \\
\hline
cmn &-- & \textcolor{lightgray}{14} & \textcolor{lightgray}{25} & \textbf{56} & \textcolor{lightgray}{18} & 113 \\
deu &\textcolor{lightgray}{17} & -- & \textcolor{lightgray}{10} & \textbf{30} & \textbf{18} & 75 \\
eng &\textcolor{lightgray}{19} & \textcolor{lightgray}{8} & -- & \textbf{57} & \textcolor{lightgray}{12} & 96 \\
rus &11 & 9 & 9 & -- & \textcolor{lightgray}{13} & 42 \\
tur &\textcolor{lightgray}{6} & 3 & \textcolor{lightgray}{13} & \textcolor{lightgray}{27} & -- & 49 \\
\hline
\multicolumn{7}{c}{\sc Llama-3.1-8B-Instruct} \\
\hline
cmn &-- & \textbf{19} & \textbf{33} & \textbf{64} & \textcolor{lightgray}{12} & 128 \\
deu &3 & -- & \textbf{21} & \textbf{70} & \textcolor{lightgray}{11} & 105 \\
eng &2 & 4 & -- & \textbf{61} & \textcolor{lightgray}{7} & 74 \\
rus &19 & 15 & 19 & -- & 11 & 64 \\
tur &\textcolor{lightgray}{11} & \textcolor{lightgray}{17} & \textcolor{lightgray}{12} & \textbf{68} & -- & 108 \\
\hline
\multicolumn{7}{c}{\sc gpt-5.2-2025-12-11} \\
\hline
cmn &-- & 7 & 4 & \textcolor{lightgray}{26} & 15 & 52 \\
deu &\textbf{61} & -- & 14 & \textbf{52} & \textbf{47} & 174 \\
eng &\textbf{73} & \textbf{42} & -- & \textbf{78} & \textbf{75} & 268 \\
rus &\textcolor{lightgray}{22} & 11 & 2 & -- & \textcolor{lightgray}{23} & 58 \\
tur &\textbf{40} & 8 & 8 & \textcolor{lightgray}{34} & -- & 90 \\
\hline
\multicolumn{7}{c}{\sc gpt-5-mini-2025-08-07} \\
\hline
cmn &-- & 17 & 10 & \textbf{67} & \textcolor{lightgray}{26} & 120 \\
deu &\textbf{48} & -- & 19 & \textbf{70} & \textcolor{lightgray}{41} & 178 \\
eng &\textbf{61} & \textbf{48} & -- & \textbf{91} & \textcolor{lightgray}{38} & 238 \\
rus &12 & 9 & 6 & -- & 16 & 43 \\
tur &\textcolor{lightgray}{38} & \textcolor{lightgray}{26} & \textcolor{lightgray}{16} & \textbf{75} & -- & 155 \\
\hline
\multicolumn{7}{c}{\sc gpt-5-nano-2025-08-07} \\
\hline
cmn &-- & \textbf{39} & \textbf{44} & \textbf{51} & 6 & 140 \\
deu &10 & -- & \textcolor{lightgray}{24} & \textbf{41} & \textcolor{lightgray}{13} & 88 \\
eng &10 & \textcolor{lightgray}{21} & -- & \textcolor{lightgray}{30} & 8 & 69 \\
rus &1 & 1 & \textcolor{lightgray}{12} & -- & 1 & 15 \\
tur &\textbf{37} & \textcolor{lightgray}{30} & \textbf{45} & \textbf{69} & -- & 181 \\
\hline
\end{tabular}
\end{minipage}
\hfill
\begin{minipage}{0.5\textwidth}
\begin{tabular}{l|rrrrr|r}
 & cmn & deu & eng & rus & tur & $\Sigma$ \\
\hline
\multicolumn{7}{c}{\sc c4ai-command-r7b-12-2024} \\
\hline
cmn &-- & \textcolor{lightgray}{15} & 12 & \textcolor{lightgray}{43} & \textcolor{lightgray}{15} & 85 \\
deu &\textcolor{lightgray}{27} & -- & 0 & \textcolor{lightgray}{48} & \textcolor{lightgray}{27} & 102 \\
eng &\textbf{49} & \textbf{29} & -- & \textcolor{lightgray}{35} & \textbf{47} & 160 \\
rus &\textcolor{lightgray}{35} & \textcolor{lightgray}{27} & \textcolor{lightgray}{39} & -- & 24 & 125 \\
tur &\textcolor{lightgray}{3} & \textcolor{lightgray}{9} & 5 & \textbf{53} & -- & 70 \\
\hline
\multicolumn{7}{c}{\sc Ministral-8B-Instruct-2410} \\
\hline
cmn &-- & \textcolor{lightgray}{16} & \textcolor{lightgray}{24} & \textbf{79} & \textcolor{lightgray}{8} & 127 \\
deu &\textcolor{lightgray}{10} & -- & \textcolor{lightgray}{10} & \textbf{65} & \textcolor{lightgray}{10} & 95 \\
eng &\textcolor{lightgray}{11} & \textcolor{lightgray}{8} & -- & \textbf{72} & \textcolor{lightgray}{8} & 99 \\
rus &4 & 8 & 10 & -- & 7 & 29 \\
tur &\textcolor{lightgray}{6} & \textcolor{lightgray}{10} & \textcolor{lightgray}{14} & \textbf{46} & -- & 76 \\
\hline
\multicolumn{7}{c}{\sc Mistral-Nemo-Instruct-2407} \\
\hline
cmn &-- & \textcolor{lightgray}{22} & \textcolor{lightgray}{24} & \textbf{61} & \textcolor{lightgray}{25} & 132 \\
deu &\textcolor{lightgray}{9} & -- & \textcolor{lightgray}{6} & \textbf{64} & \textcolor{lightgray}{7} & 86 \\
eng &\textcolor{lightgray}{8} & \textcolor{lightgray}{7} & -- & \textbf{57} & \textcolor{lightgray}{17} & 89 \\
rus &5 & 9 & 9 & -- & 7 & 30 \\
tur &\textcolor{lightgray}{7} & \textcolor{lightgray}{7} & \textcolor{lightgray}{13} & \textbf{63} & -- & 90 \\
\hline
\multicolumn{7}{c}{\sc GLM-4-9B-0414} \\
\hline
cmn &-- & \textcolor{lightgray}{14} & \textcolor{lightgray}{28} & \textbf{69} & \textcolor{lightgray}{20} & 131 \\
deu &\textcolor{lightgray}{22} & -- & \textcolor{lightgray}{18} & \textbf{43} & \textcolor{lightgray}{12} & 95 \\
eng &\textcolor{lightgray}{12} & \textcolor{lightgray}{9} & -- & \textbf{47} & \textcolor{lightgray}{16} & 84 \\
rus &8 & 3 & 9 & -- & 4 & 24 \\
tur &\textcolor{lightgray}{9} & \textcolor{lightgray}{14} & \textcolor{lightgray}{8} & \textbf{39} & -- & 70 \\
\hline
\multicolumn{7}{c}{\sc Qwen3-4B} \\
\hline
cmn &-- & \textcolor{lightgray}{15} & \textcolor{lightgray}{18} & \textbf{64} & \textcolor{lightgray}{12} & 109 \\
deu &\textcolor{lightgray}{23} & -- & \textcolor{lightgray}{18} & \textbf{63} & \textcolor{lightgray}{23} & 127 \\
eng &\textcolor{lightgray}{20} & \textcolor{lightgray}{5} & -- & \textbf{57} & \textcolor{lightgray}{18} & 100 \\
rus &3 & 1 & 2 & -- & 8 & 14 \\
tur &\textcolor{lightgray}{22} & \textcolor{lightgray}{15} & \textcolor{lightgray}{33} & \textbf{77} & -- & 147 \\
\hline
\multicolumn{7}{c}{\sc Yi-1.5-9B-32K} \\
\hline
cmn &-- & \textcolor{lightgray}{28} & \textcolor{lightgray}{20} & \textbf{67} & \textbf{34} & 149 \\
deu &\textcolor{lightgray}{9} & -- & \textcolor{lightgray}{10} & \textbf{46} & \textcolor{lightgray}{11} & 76 \\
eng &\textcolor{lightgray}{14} & \textcolor{lightgray}{9} & -- & \textbf{55} & \textbf{34} & 112 \\
rus &0 & 1 & 4 & -- & 1 & 6 \\
tur &4 & \textcolor{lightgray}{10} & 12 & \textbf{64} & -- & 90 \\
\hline
\end{tabular}
\end{minipage}
\hfill
\caption{Total number of times, over all prompt languages, where language $L_1$ (row) is chosen over $L_2$ (column) in \emph{both} orders of a contrastive pair. Bold: row language wins over column language. Light gray: non-significant difference (Bonferroni-corrected $p < 0.05$). Haystack size is 5000.}
\label{tab:pairs5k}
\end{table*}
\begin{table*}

\begin{minipage}{0.5\textwidth}
\begin{tabular}{l|rrrrr|r}
 & cmn & deu & eng & rus & tur & $\Sigma$ \\
\hline
\multicolumn{7}{c}{\sc gemma-3-27b-it} \\
\hline
cmn &-- & \textcolor{lightgray}{10} & \textcolor{lightgray}{14} & \textbf{44} & \textcolor{lightgray}{15} & 83 \\
deu &\textcolor{lightgray}{8} & -- & \textcolor{lightgray}{13} & \textcolor{lightgray}{26} & \textcolor{lightgray}{9} & 56 \\
eng &\textcolor{lightgray}{8} & \textcolor{lightgray}{8} & -- & \textbf{39} & \textcolor{lightgray}{6} & 61 \\
rus &7 & \textcolor{lightgray}{8} & 5 & -- & \textcolor{lightgray}{9} & 29 \\
tur &\textcolor{lightgray}{18} & \textcolor{lightgray}{8} & \textcolor{lightgray}{13} & \textcolor{lightgray}{26} & -- & 65 \\
\hline
\multicolumn{7}{c}{\sc gemma-3-4b-it} \\
\hline
cmn &-- & \textcolor{lightgray}{8} & \textcolor{lightgray}{15} & \textbf{47} & \textcolor{lightgray}{9} & 79 \\
deu &\textcolor{lightgray}{8} & -- & \textcolor{lightgray}{12} & \textbf{31} & \textbf{17} & 68 \\
eng &\textcolor{lightgray}{9} & \textcolor{lightgray}{6} & -- & \textbf{59} & \textcolor{lightgray}{22} & 96 \\
rus &5 & 3 & 4 & -- & 4 & 16 \\
tur &\textcolor{lightgray}{3} & 0 & \textcolor{lightgray}{9} & \textbf{32} & -- & 44 \\
\hline
\multicolumn{7}{c}{\sc Llama-3.1-8B-Instruct} \\
\hline
cmn &-- & \textcolor{lightgray}{10} & \textcolor{lightgray}{19} & \textbf{69} & \textcolor{lightgray}{4} & 102 \\
deu &\textcolor{lightgray}{13} & -- & \textcolor{lightgray}{15} & \textbf{67} & \textcolor{lightgray}{4} & 99 \\
eng &\textcolor{lightgray}{7} & \textcolor{lightgray}{2} & -- & \textbf{59} & 2 & 70 \\
rus &20 & 12 & 22 & -- & 12 & 66 \\
tur &\textcolor{lightgray}{6} & \textcolor{lightgray}{8} & \textbf{20} & \textbf{81} & -- & 115 \\
\hline
\multicolumn{7}{c}{\sc gpt-5.2-2025-12-11} \\
\hline
cmn &-- & 8 & 2 & \textcolor{lightgray}{21} & \textcolor{lightgray}{18} & 49 \\
deu &\textbf{61} & -- & 5 & \textbf{51} & \textbf{40} & 157 \\
eng &\textbf{77} & \textbf{47} & -- & \textbf{78} & \textbf{57} & 259 \\
rus &\textcolor{lightgray}{19} & 7 & 3 & -- & \textcolor{lightgray}{20} & 49 \\
tur &\textcolor{lightgray}{33} & 13 & 10 & \textcolor{lightgray}{35} & -- & 91 \\
\hline
\multicolumn{7}{c}{\sc gpt-5-mini-2025-08-07} \\
\hline
cmn &-- & \textcolor{lightgray}{15} & 10 & \textbf{65} & \textcolor{lightgray}{18} & 108 \\
deu &\textcolor{lightgray}{37} & -- & \textcolor{lightgray}{18} & \textbf{78} & \textcolor{lightgray}{42} & 175 \\
eng &\textbf{57} & \textcolor{lightgray}{34} & -- & \textbf{93} & \textbf{46} & 230 \\
rus &13 & 4 & 5 & -- & 17 & 39 \\
tur &\textcolor{lightgray}{26} & \textcolor{lightgray}{19} & 15 & \textbf{58} & -- & 118 \\
\hline
\multicolumn{7}{c}{\sc gpt-5-nano-2025-08-07} \\
\hline
cmn &-- & \textbf{27} & \textbf{57} & \textbf{63} & \textcolor{lightgray}{13} & 160 \\
deu &7 & -- & \textcolor{lightgray}{26} & \textbf{30} & 4 & 67 \\
eng &11 & \textcolor{lightgray}{17} & -- & \textbf{39} & 10 & 77 \\
rus &2 & 5 & 5 & -- & 0 & 12 \\
tur &\textcolor{lightgray}{28} & \textbf{35} & \textbf{57} & \textbf{70} & -- & 190 \\
\hline
\end{tabular}
\end{minipage}
\hfill
\begin{minipage}{0.5\textwidth}
\begin{tabular}{l|rrrrr|r}
 & cmn & deu & eng & rus & tur & $\Sigma$ \\
\hline
\multicolumn{7}{c}{\sc c4ai-command-r7b-12-2024} \\
\hline
cmn &-- & \textcolor{lightgray}{11} & \textcolor{lightgray}{14} & \textcolor{lightgray}{41} & \textcolor{lightgray}{17} & 83 \\
deu &\textcolor{lightgray}{20} & -- & 1 & \textcolor{lightgray}{44} & \textcolor{lightgray}{24} & 89 \\
eng &\textcolor{lightgray}{30} & \textbf{38} & -- & \textcolor{lightgray}{41} & \textbf{42} & 151 \\
rus &\textcolor{lightgray}{25} & \textcolor{lightgray}{24} & \textcolor{lightgray}{38} & -- & 22 & 109 \\
tur &\textcolor{lightgray}{10} & \textcolor{lightgray}{7} & 8 & \textbf{57} & -- & 82 \\
\hline
\multicolumn{7}{c}{\sc Ministral-8B-Instruct-2410} \\
\hline
cmn &-- & \textcolor{lightgray}{11} & \textcolor{lightgray}{24} & \textbf{68} & \textcolor{lightgray}{9} & 112 \\
deu &\textcolor{lightgray}{17} & -- & \textcolor{lightgray}{7} & \textbf{66} & \textcolor{lightgray}{15} & 105 \\
eng &\textcolor{lightgray}{13} & \textcolor{lightgray}{2} & -- & \textbf{58} & \textcolor{lightgray}{13} & 86 \\
rus &9 & 8 & 11 & -- & 10 & 38 \\
tur &\textcolor{lightgray}{13} & \textcolor{lightgray}{8} & \textcolor{lightgray}{15} & \textbf{53} & -- & 89 \\
\hline
\multicolumn{7}{c}{\sc Mistral-Nemo-Instruct-2407} \\
\hline
cmn &-- & \textbf{31} & \textcolor{lightgray}{26} & \textbf{69} & \textcolor{lightgray}{26} & 152 \\
deu &10 & -- & \textcolor{lightgray}{4} & \textbf{47} & \textcolor{lightgray}{18} & 79 \\
eng &\textcolor{lightgray}{17} & \textcolor{lightgray}{8} & -- & \textcolor{lightgray}{38} & \textcolor{lightgray}{24} & 87 \\
rus &9 & 14 & \textcolor{lightgray}{16} & -- & 7 & 46 \\
tur &\textcolor{lightgray}{12} & \textcolor{lightgray}{18} & \textcolor{lightgray}{17} & \textbf{57} & -- & 104 \\
\hline
\multicolumn{7}{c}{\sc GLM-4-9B-0414} \\
\hline
cmn &-- & \textcolor{lightgray}{30} & \textbf{41} & \textbf{58} & \textcolor{lightgray}{15} & 144 \\
deu &\textcolor{lightgray}{15} & -- & \textbf{28} & \textbf{30} & \textcolor{lightgray}{13} & 86 \\
eng &4 & 6 & -- & \textbf{41} & \textcolor{lightgray}{7} & 58 \\
rus &1 & 4 & 11 & -- & 4 & 20 \\
tur &\textcolor{lightgray}{5} & \textcolor{lightgray}{12} & \textcolor{lightgray}{20} & \textbf{44} & -- & 81 \\
\hline
\multicolumn{7}{c}{\sc Qwen3-4B} \\
\hline
cmn &-- & \textcolor{lightgray}{14} & \textcolor{lightgray}{13} & \textbf{68} & \textcolor{lightgray}{9} & 104 \\
deu &\textcolor{lightgray}{29} & -- & \textcolor{lightgray}{20} & \textbf{59} & \textcolor{lightgray}{11} & 119 \\
eng &\textcolor{lightgray}{14} & \textcolor{lightgray}{9} & -- & \textbf{61} & \textcolor{lightgray}{23} & 107 \\
rus &5 & 0 & 1 & -- & 1 & 7 \\
tur &\textcolor{lightgray}{27} & \textcolor{lightgray}{23} & \textcolor{lightgray}{27} & \textbf{91} & -- & 168 \\
\hline
\multicolumn{7}{c}{\sc Yi-1.5-9B-32K} \\
\hline
cmn &-- & \textcolor{lightgray}{16} & \textcolor{lightgray}{13} & \textbf{57} & \textbf{47} & 133 \\
deu &\textcolor{lightgray}{13} & -- & \textcolor{lightgray}{4} & \textbf{33} & \textcolor{lightgray}{4} & 54 \\
eng &\textcolor{lightgray}{8} & \textcolor{lightgray}{2} & -- & \textbf{42} & \textcolor{lightgray}{19} & 71 \\
rus &1 & 3 & 4 & -- & 1 & 9 \\
tur &1 & \textcolor{lightgray}{14} & \textcolor{lightgray}{11} & \textbf{45} & -- & 71 \\
\hline
\end{tabular}
\end{minipage}
\hfill
\caption{Total number of times, over all prompt languages, where language $L_1$ (row) is chosen over $L_2$ (column) in \emph{both} orders of a contrastive pair. Bold: row language wins over column language. Light gray: non-significant difference (Bonferroni-corrected $p < 0.05$). Haystack size is 10000.}
\label{tab:pairs10k}
\end{table*}

\section{Non-conflicting monolingual retrieval}
\label{sec:nonconflicting-monolingual}

\Fref{tab:nonconflicting-monolingual} shows the retrieval rates for the 12 attempts per needle language and prompt language in the easiest of our settings, non-conflicting monolingual haystacks. All models are able to achieve a perfect result in at least one language (typically English and/or Chinese), although some models struggle with certain needle and/or prompt languages.

\begin{table*}
\begin{minipage}{0.5\textwidth}
\begin{tabular}{l|rrrrr}
Prompt & cmn & deu & eng & rus & tur \\
\hline
\multicolumn{6}{c}{\sc gemma-3-27b-it} \\
\hline
cmn &10 & 10 & 12 & 10 & 12 \\
deu &12 & 12 & 12 & 11 & 12 \\
eng &12 & 11 & 12 & 12 & 12 \\
rus &11 & 11 & 12 & 12 & 10 \\
tur &11 & 11 & 12 & 9 & 12 \\
\hline
\multicolumn{6}{c}{\sc gemma-3-4b-it} \\
\hline
cmn &9 & 5 & 9 & 4 & 9 \\
deu &10 & 7 & 11 & 6 & 12 \\
eng &11 & 10 & 12 & 7 & 10 \\
rus &8 & 9 & 10 & 10 & 11 \\
tur &9 & 6 & 8 & 5 & 9 \\
\hline
\multicolumn{6}{c}{\sc Llama-3.1-8B-Instruct} \\
\hline
cmn &12 & 12 & 12 & 12 & 12 \\
deu &12 & 12 & 12 & 10 & 12 \\
eng &12 & 12 & 12 & 11 & 12 \\
rus &12 & 12 & 11 & 12 & 12 \\
tur &12 & 12 & 12 & 12 & 12 \\
\hline
\multicolumn{6}{c}{\sc gpt-5.2-2025-12-11} \\
\hline
cmn &12 & 12 & 12 & 12 & 12 \\
deu &12 & 12 & 12 & 12 & 12 \\
eng &12 & 12 & 12 & 12 & 12 \\
rus &12 & 12 & 12 & 12 & 12 \\
tur &12 & 12 & 12 & 12 & 12 \\
\hline
\multicolumn{6}{c}{\sc gpt-5-mini-2025-08-07} \\
\hline
cmn &12 & 12 & 12 & 12 & 12 \\
deu &12 & 12 & 12 & 12 & 12 \\
eng &12 & 12 & 12 & 12 & 12 \\
rus &12 & 12 & 12 & 12 & 12 \\
tur &12 & 12 & 12 & 12 & 12 \\
\hline
\multicolumn{6}{c}{\sc gpt-5-nano-2025-08-07} \\
\hline
cmn &12 & 12 & 12 & 12 & 12 \\
deu &12 & 12 & 12 & 12 & 12 \\
eng &12 & 12 & 12 & 12 & 12 \\
rus &12 & 12 & 12 & 12 & 12 \\
tur &12 & 12 & 12 & 12 & 12 \\
\hline
\end{tabular}
\end{minipage}
\hfill
\begin{minipage}{0.5\textwidth}
\begin{tabular}{l|rrrrr}
Prompt & cmn & deu & eng & rus & tur \\
\hline
\multicolumn{6}{c}{\sc c4ai-command-r7b-12-2024} \\
\hline
cmn &12 & 12 & 12 & 11 & 12 \\
deu &12 & 12 & 12 & 12 & 12 \\
eng &12 & 12 & 12 & 12 & 12 \\
rus &12 & 12 & 12 & 12 & 12 \\
tur &12 & 11 & 12 & 9 & 12 \\
\hline
\multicolumn{6}{c}{\sc Ministral-8B-Instruct-2410} \\
\hline
cmn &12 & 12 & 12 & 12 & 12 \\
deu &12 & 12 & 12 & 12 & 12 \\
eng &12 & 12 & 12 & 12 & 12 \\
rus &12 & 12 & 12 & 12 & 12 \\
tur &12 & 12 & 12 & 12 & 12 \\
\hline
\multicolumn{6}{c}{\sc Mistral-Nemo-Instruct-2407} \\
\hline
cmn &7 & 5 & 8 & 1 & 5 \\
deu &10 & 8 & 11 & 2 & 3 \\
eng &9 & 9 & 12 & 4 & 8 \\
rus &7 & 9 & 10 & 9 & 4 \\
tur &8 & 5 & 10 & 1 & 9 \\
\hline
\multicolumn{6}{c}{\sc GLM-4-9B-0414} \\
\hline
cmn &10 & 10 & 12 & 10 & 10 \\
deu &12 & 12 & 12 & 6 & 0 \\
eng &11 & 11 & 12 & 8 & 11 \\
rus &12 & 12 & 12 & 8 & 8 \\
tur &10 & 1 & 8 & 4 & 2 \\
\hline
\multicolumn{6}{c}{\sc Qwen3-4B} \\
\hline
cmn &12 & 12 & 12 & 12 & 12 \\
deu &12 & 12 & 12 & 7 & 12 \\
eng &12 & 12 & 12 & 6 & 12 \\
rus &12 & 12 & 12 & 12 & 12 \\
tur &12 & 12 & 12 & 8 & 12 \\
\hline
\multicolumn{6}{c}{\sc Yi-1.5-9B-32K} \\
\hline
cmn &12 & 4 & 12 & 1 & 3 \\
deu &12 & 6 & 9 & 0 & 5 \\
eng &12 & 6 & 12 & 0 & 5 \\
rus &12 & 2 & 10 & 1 & 4 \\
tur &12 & 4 & 10 & 0 & 4 \\
\hline
\end{tabular}
\end{minipage}
\caption{Total number of times a given \textbf{non-conflicting} was successfully retrieved in a \textbf{monolingual} haystack, by prompt language. Haystack size is 25~000.}
\label{tab:nonconflicting-monolingual}
\end{table*}

\end{document}